%% file: main.tex
\documentclass[10pt,twocolumn,letterpaper]{article}

\usepackage{cvpr}              

\usepackage{algorithm}       
\usepackage{algpseudocode} 
\usepackage{subcaption}
\usepackage{multirow}
\usepackage{placeins}
\usepackage{amsmath,amssymb,amsthm}
\usepackage{booktabs, tabularx, makecell}
\usepackage{comment}
\newcolumntype{Y}{>{\raggedright\arraybackslash}X} 
\newcommand{\rob}{Rob-C-Bar}
\newcommand{\flip}{Flip}
\newcommand{\robc}{Rob C-bar}
\newcommand{\pmerr}[2]{#1\,\text{\tiny$\pm$ #2}}
\input{preamble}
\definecolor{cvprblue}{rgb}{0.21,0.49,0.74}
\usepackage[pagebackref,breaklinks,colorlinks,allcolors=cvprblue]{hyperref}

\def\paperID{19} 
\def\confName{CVPR}
\def\confYear{2026}

\title{DART: A Server-side Plug-in for Resource-efficient Robust Federated Learning}

\author{Omar Bekdache and Naresh Shanbhag\\
University of Illinois Urbana-Champaign\\
1308 W. Main St. Urbana, IL 61801\\
{\tt\small omarb3@illinois.edu, shanbhag@illinois.edu}
}

\newcommand{\x}{\mathbf{x}} 
\newcommand{\inx}{\mathbf{x}_\text{in}} 
\newcommand{\testx}{\mathbf{x}_\text{test}} 
\newcommand{\corrx}{\mathbf{x}_\text{corr}} 

\newcommand{\w}{\mathbf{w}} 
\newcommand{\y}{\mathbf{y}} 

\newcommand{\inD}{\mathcal{D}_\text{in}} 
\newcommand{\outD}{\mathcal{D}_\text{out}} 
\newcommand{\corrD}{\mathcal{D}_\text{corr}} 

\newcommand{\A}{\mathcal{A}} 
\newcommand{\R}{\mathcal{R}} 
\newcommand{\clnA}{\mathcal{A}_\text{cln}} 
\newcommand{\robA}{\mathcal{A}_\text{rob}} 
\newcommand{\avgA}{\mathcal{A}_\text{avg}} 

\newcommand{\DARTw}{\mathbf{w}_\text{rob}} 

\newcommand{\DARTloss}{\mathcal{L}_\text{DART}} 
\newcommand{\closs}{\mathcal{L}_\text{c}} 
\newcommand{\dloss}{\mathcal{L}_\text{d}} 

\newcommand{\robT}{T_\text{DART}} 

\newcommand{\batch}{\mathbf{b}} 

\newcommand{\DARTD}{D_\text{tr}} 
\newcommand{\valD}{D_\text{val}} 

\newcommand{\DARTx}{\mathbf{x}_\text{tr}} 
\newcommand{\augfirstx}{\mathbf{x}_\text{aug1}} 
\newcommand{\augsecondx}{\mathbf{x}_\text{aug2}} 

\newcommand{\DARTp}{\mathbf{p}_\text{s}} 
\newcommand{\augfirstp}{\mathbf{p}_\text{s,aug1}} 
\newcommand{\augsecondp}{\mathbf{p}_\text{s,aug2}} 

\newcommand{\teacherp}{\mathbf{p}_\text{t}} 
\newcommand{\studentp}{\mathbf{p}_\text{s}} 

\newcommand{\valT}{T_\text{val}} 
\newcommand{\maxT}{T_\text{max}} 

\newcommand{\trainTime}{\mathcal{T}_{\text{tr}}} %
\newcommand{\trainEnergy}{\mathcal{E}_{\text{tr}}} %
\newcommand{\trainMemory}{\mathcal{M}_{\text{tr}}} %

\newcommand{\DARTeta}{\eta_\text{s}} 

\newcommand{\augx}{\mathbf{x}_\text{aug}} 

\newtheorem{theorem}{Theorem}
\newtheorem{lemma}{Lemma}
\newtheorem*{theorem*}{Theorem}  

\begin{document}
\maketitle
\input{sec/0_abstract}    
\input{sec/1_intro_new}
\input{sec/2_background_and_related}

\input{sec/4_notation}
\input{sec/5_method}

\input{sec/6_results}

\input{sec/7_conclusion}
\FloatBarrier
{
    \small
    \bibliographystyle{ieeenat_fullname}
    \bibliography{main}
}

\input{sec/X_suppl}

\end{document}

%% file: sec/0_abstract.tex
\begin{abstract}
Federated learning (FL) emerged as a popular distributed algorithm to train machine learning models on edge devices while preserving data privacy. However, FL systems face challenges due to client-side computational constraints and from a lack of robustness to naturally occurring common corruptions such as noise, blur, and weather effects. Existing robust training methods are computationally expensive and unsuitable for resource-constrained clients. We propose a novel data-agnostic robust training (DART) plug-in that can be deployed in any FL system to enhance robustness at \emph{zero} client overhead. DART operates at the server-side and does not require private data access, ensuring seamless integration in existing FL systems. Extensive experiments showcase DART's ability to enhance robustness of state-of-the-art FL systems, establishing it as a practical and scalable solution for real-world robust FL deployment.
\end{abstract}

%% file: sec/1_intro_new.tex
\section{Introduction}
\label{sec:Introduction}
\begin{figure*}[t]
\begin{center}
\includegraphics[width=\textwidth]{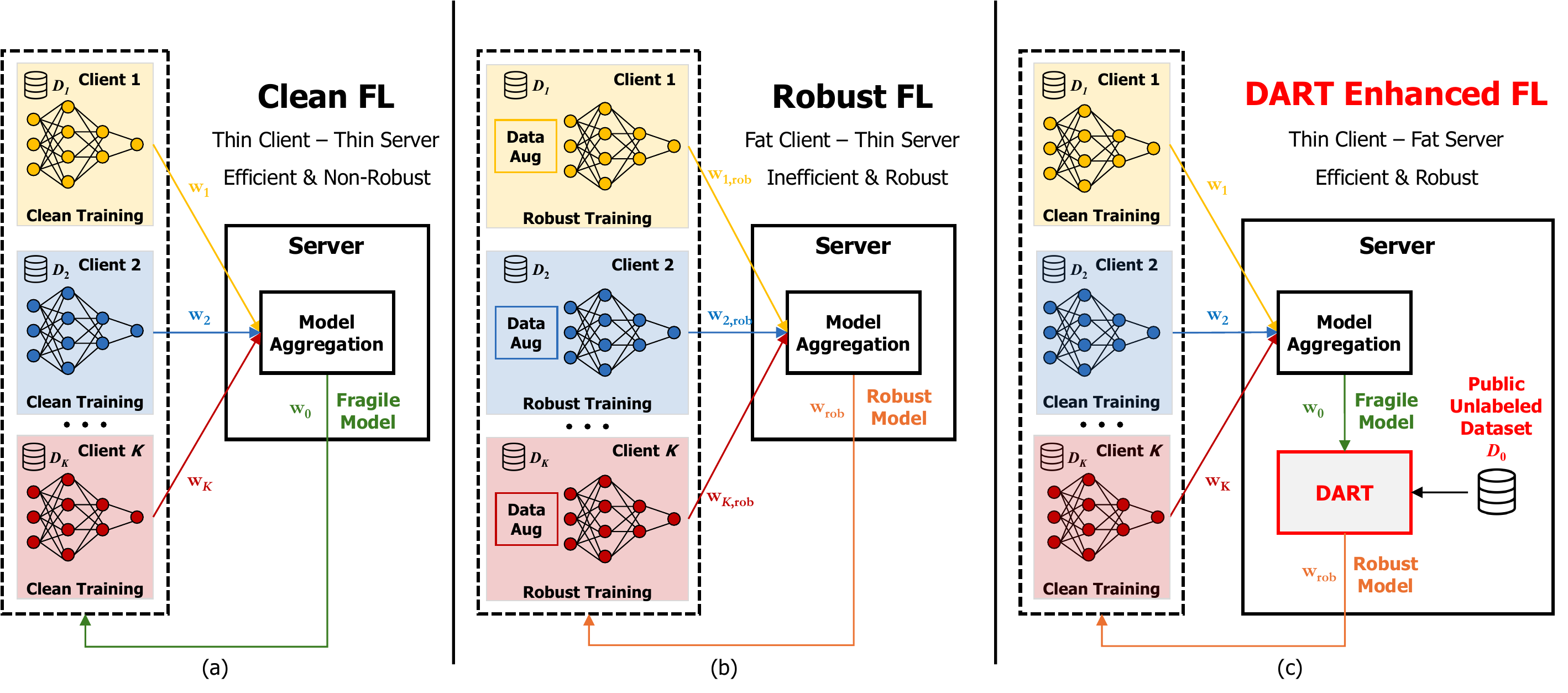}

    \caption{Clients in conventional Clean FL (a) employ local data to train while the server aggregates the model to generate a fragile model. Robust FL (b) employs computationally intensive data augmentation and robust training on the clients while the server aggregates as in Clean FL. Our proposed DART-enhanced FL method (c) employs low-cost clean training on the clients and the data-agnostic robustness training (DART) method on the server to train a robust model, exploiting the asymmetry in the computational resources (time and energy) between the clients and the server.}
    \label{fig:methods}
\end{center}
\vskip -0.3in
\end{figure*}

Federated learning (FL) has emerged as a promising distributed learning paradigm that enables model training across decentralized data sources while preserving data privacy \cite{mcmahan2017communication, yang2019federated} (Fig.~\ref{fig:methods}a). 
This paradigm is especially beneficial for deep learning, where model performance scales with data volume \cite{hestness2017deep} and privacy constraints often restrict centralized training. Federated learning has been explored in various domains, including healthcare \cite{antunes2022federated}, finance \cite{wen2023survey}, computer vision \cite{shenaj2023federated}, and commercial settings \cite{yang2018applied}.

Conventionally, FL methods place the entire computational burden of training on the clients, while the server implements the relatively simple task of model aggregation. While this workload allocation ensures data privacy, it is completely mismatched to the computational resources available at the clients and the server. Clients, typically mobile or embedded devices, are heavily resource-constrained~\cite{han2015learning, chen2023efficient}, whereas the server is comparatively resource-rich. 

This asymmetry in compute resources is amplified significantly when training models that are robust to out-of-distribution (OOD) inputs, such as adversarial~\cite{zizzo2020fat, chen2022calfat, qiao2024logit} and common corruption robustness~\cite{fang2023robust} (Fig.~\ref{fig:methods}b).

In this work, we focus on common corruption robustness, which, despite its importance for achieving reliable FL performance in real-world edge environments, has received limited attention by the FL community. While several methods improve robustness to common corruptions~\cite{hendrycks2019augmix, modas2022prime, vaish2024fourier}, these approaches are often impractical under client-side resource constraints. Common corruption robust FL training faces challenges from two main factors: (1) robust training is entirely executed locally on clients to preserve data privacy~\cite{fang2023robust}, and (2) common corruption robustness objectives substantially increase computational cost~\cite{vaish2024fourier}. These demands render robust FL impractical for deployment on edge devices, as further discussed in Section~\ref{sec:background_related}. Given that such devices often run multiple concurrent workloads, reducing the client-side cost of robust FL is a key design goal. Despite its practical significance, common corruption robust methods do not address client-side resource constraints.

In this work, we introduce DART, a \textbf{\underline{D}}ata-\textbf{\underline{A}}gnostic \textbf{\underline{R}}obust \textbf{\underline{T}}raining framework designed as a universal server-side plug-in to enhance the common corruption robustness of any FL system. A DART-enhanced FL system simultaneously \emph{ensures data privacy and zero client overhead} on computation, memory, time, and energy (Fig.~\ref{fig:methods}c). In such a system, clients train using private data while all robustness enhancements are realized on the server employing a public dataset. Central to DART is the idea of decoupling utility maximization from robustness enhancement. While clients perform the comparatively cheap clean training to improve utility, the server is fully responsible for enhancing robustness. DART adopts a teacher–student framework that integrates a data augmentation scheme, a utility-preserving loss, and a consistency-enhancing objective in order to improve robustness without sacrificing accuracy. 

We summarize our key contributions as follows:
\begin{enumerate}
    \item We propose DART, a data-agnostic robust training framework that enhances the corruption robustness of any machine learning model without requiring access to the target data distribution.
    \item We integrate DART with multiple popular FL algorithms, and we empirically show that DART achieves favorable utility–robustness trade-offs across diverse FL algorithms with zero client-side overhead in computation, memory, time, and energy.
    \item We derive a theoretical upper bound on the utility risk of DART-enhanced models and empirically show that DART preserves strong utility and robustness under distribution mismatch.
\end{enumerate}

%% file: sec/2_background_and_related.tex
\section{Background and Related Works}
\label{sec:background_related}

\paragraph{Federated Learning.}~\citet{mcmahan2017communication} introduced the first FL algorithm, FedAvg, where clients independently train local models on their private data and transmit model parameters to the server. The server aggregates these parameters and returns the updated global model back to the clients for further local training. The area quickly grew to include more recent works that study the impact of data heterogeneity, where client datasets follow different distributions, and propose methods to mitigate the resulting performance drop~\cite{li2020federated, karimireddy2020scaffold, kim2022multi, zhang2022federated, wang2024dfrd}. Another area of research is model-heterogeneous federated learning, where clients possess different model architectures while the server still performs aggregation~\cite{li2019fedmd, diao2020heterofl, yi2023fedgh, yao2025fedhm}. Energy efficiency has also emerged as a significant area of interest in FL research~\cite{khowaja2021toward, kim2021autofl, yang2020energy}. 

Server-side training has been investigated in FL to address the scarcity of labeled client data~\cite{wang2022enhancing} and to alleviate data and model heterogeneity~\cite{wang2023dfrd}. However, decoupling model utility from robustness by shifting robustness related computation to the resource-rich server has not been explored. This design is essential for enabling common corruption robust FL models, particularly when clients impose time, energy and memory constraints.

\paragraph{Common Corruptions Robustness.}
The deployment of deep neural networks, including those trained under FL, is often constrained by their lack of robustness. In edge-device vision applications, data captured on-device is frequently affected by natural common corruptions such as adverse weather conditions, sensor defects, or storage and compression artifacts~\cite{hendrycks2019benchmarking}, leading to substantial performance degradation. 

The most common approaches for training models robust to common corruptions augment the original training dataset by combining randomized image transformations and incorporating tailored loss functions~\cite{hendrycks2019augmix, modas2022prime}. For instance,~\cite{hendrycks2021many} utilizes an image-to-image network with randomly distorted parameters to generate augmented samples for training, while~\cite{vaish2024fourier} targets augmentations in the frequency domain and incorporates separate batch normalization layers and two loss terms reflecting original and augmented images. While these methods improve robustness, they come at a substantial computational cost. 
\textbf{In contrast, our work employs a publicly available unlabeled dataset as a proxy on the server, thereby offloading robust training and reducing the workload burden on the clients.} 

\begin{table}[t]
\centering
\caption{Client-side resource consumption relative to FedAvg, measured on NVIDIA Jetson Orin Nano using ResNet-18 with batch size 64 and input dimension $32\times32\times3$. Robust FL methods exhibit significantly higher resource demands.}
\resizebox{\columnwidth}{!}{%
\begin{tabular}{lccccc}
\toprule
 & \multicolumn{2}{c}{Clean FL} & \multicolumn{3}{c}{Robust FL} \\
 \cmidrule(lr){2-3}
 \cmidrule(lr){4-6}
 & FedAvg & FedProx & FedAFA & FedAugMix & FedPrime \\
\midrule
FLOPS & $\times1.00$ & $\times1.00$ & $\times2.00$ & $\times3.00$ & $\times3.00$ \\
Memory & $\times1.00$ & $\times1.30$ & $\times1.70$ & $\times2.57$ & $\times2.57$ \\
Time & $\times1.00$ & $\times1.06$ & $\times1.94$ & $\times2.64$ & $\times21.46$ \\
Energy  & $\times1.00$ & $\times1.07$ & $\times1.93$ & $\times2.90$ & $\times5.73$  \\
\bottomrule
\end{tabular}%
}

\label{tab:fed_ratios_times}
\end{table}

\paragraph{Common Corruptions Robust FL.} FL under common corruptions has been investigated by~\cite{li2022auto, stripelis2022performance} to address the problem of malicious clients injecting corrupted data during training to degrade model performance. \textbf{In contrast, we solve a different problem: improving robustness by generalizing to corruptions unseen at training time.} The method in~\cite{fang2023robust} (FedAugMix) ignores client-side computational resource constraints to improve model robustness by having clients implement the popular robust training algorithm AugMix~\cite{hendrycks2019augmix}. 

However, clients in FL are unable to afford common corruption robust training due to excessive time and energy requirements. When conventional robustness methods are applied, client devices incur substantial computational and memory burdens. Our measurements on the NVIDIA Jetson Orin Nano (Table~\ref{tab:fed_ratios_times}) show that robust FL typically requires $2\text{-}3\times$ more FLOPs and $1.7\text{-}2.6\times$ more memory than clean training, leading to $1.9\text{-}21.5\times$ longer training time and $1.9\text{-}5.7\times$ higher energy consumption, thereby making current robust FL methods unsuitable for edge deployment. The memory overhead is particularly problematic, as it cannot be mitigated even with fewer epochs. Reducing client-side cost is therefore essential for realizing practical, corruption robust FL. \textbf{Our work is the first to train common corruption robust FL models under resource constraints by transferring robust training entirely to the server.}

\paragraph{Robustness Distillation.} Prior work has explored improving model robustness through adversarial training and robust knowledge distillation. In particular, Adversarially Robust Distillation (ARD)~\cite{goldblum2020adversarially} transfers robustness from a large adversarially trained teacher to a smaller student model by aligning the student’s predictions with the teacher’s outputs under adversarial perturbations. Similarly, Robust Soft Label Adversarial Distillation (RSLAD)~\cite{zi2021revisiting} improves this process by using the teacher’s soft predictions as supervision, enabling the student to better approximate the robust behavior of the teacher.

However, these approaches typically assume the availability of a pretrained robust teacher model, are primarily motivated by model compression, and focus specifically on adversarial robustness. \textbf{In contrast, DART targets robustness to common corruptions, does not require a pretrained robust model, and is not designed for compression.} Instead, it aims to improve robustness on the server-side to exploit the client-server resource asymmetry and to enable a more balanced workload distribution.

%% file: sec/4_notation.tex
\section{Notation}
\label{sec:not_and_prel}

\paragraph{Setup.} Let $f_\w: \mathbb{R}^d \rightarrow [C]$ be a $C$-ary classifier, parametrized by weights $\w$. $f_\w(\x) \in \mathbb{R}^C$ is used to denote the soft output of classifier $f_\w$ with parameters $\w$ and input $\x$. The FL setup consists of $K$ clients and one central server. Each client $c_k \in \{ c_1, c_2, \dots, c_{K} \}$ owns a private labeled dataset $D_k = \{(\x_i^k, \y_i^k)\}_{i=1}^{N_k}$ consisting of $N_k$ samples with $D_k \sim \inD$. The server is equipped with a public unlabeled dataset $D_0 = \{\x_i^0\}_{i=1}^{N_0}$ comprising $N_0$ samples, which follows a different distribution, i.e., $D_0 \sim \outD$. 

\paragraph{Training.} Each client $c_k$ trains a classifier $f_\w$ on its respective dataset $D_k$. These parameters are periodically synchronized by the server during the global update phase, resulting in a global model $f_{\w_0}$. Synchronization occurs a total of $G$ times, once every $E$ epochs of local training.  

\paragraph{Testing.} At test-time, client $c_k$ encounters images that are "clean" ($\sim \inD$) and "corrupt" ($\sim \corrD$). We define the classification accuracy as $\A = \Pr (\hat \y = \y)$, where $\hat \y = f_{\w_0}(\testx)$. For samples $(\testx, \mathbf{y}) \sim \inD$, the accuracy is referred to as clean accuracy $\clnA$ (utility), while for samples $(\testx, \y) \sim \corrD$, the accuracy is termed robust accuracy $\robA$ (robustness). We define the clean risk as $\R = \Pr (\hat \y \ne \y)$.
The sample $(\corrx, \y) \sim \corrD$ is generated from $(\inx, \y) \sim \inD$ using the transformation $\corrx=\kappa(\inx, s)$, where $\kappa$ is a corruption filter and $s$ is the severity level. 

%% file: sec/5_method.tex
\section{Proposed Method}
\label{sec:proposed_method}
\label{sec:method}
\begin{figure*}[t]
\begin{center}
    \includegraphics[width=\textwidth]{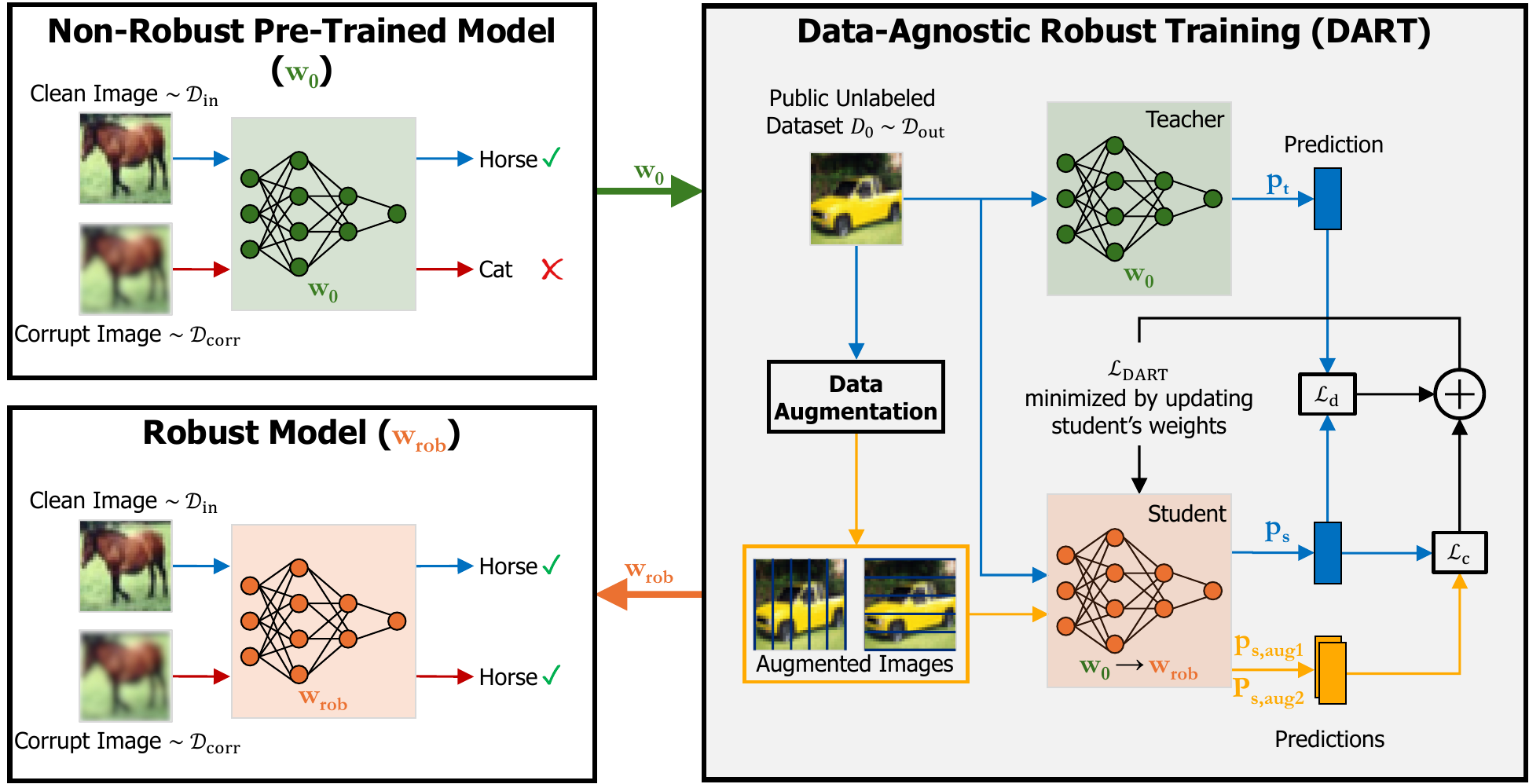}
    \caption{The proposed Data-Agnostic Robustness Training (DART) server-side plug-in. DART initializes both the teacher and student model parameters with model $f_{\w_0}$ pre-trained on clean images and then 
    minimizes $\DARTloss$~\eqref{eq:DARTloss} by updating the student model. The resulting model $f_{\DARTw}$ achieves enhanced robustness with similar clean accuracy compared to $f_{\w_0}$ with zero robust training overhead on the clients.}
    \label{fig:DART_DIAGRAM}
\end{center}
\vskip -0.3in
\end{figure*}

Our proposed approach, DART, is a server-side plug-in designed for seamless integration into existing FL frameworks. It improves system robustness without introducing additional computational overhead on clients. \textbf{A key challenge addressed by DART is achieving robustness without access to private client data, diverging from standard robust training methods that rely on direct access to the underlying data distribution.} We first present a high-level overview of a FL system augmented with DART, followed by a detailed description of its core components.

\subsection{Overall System}
\label{sec:system}
In a typical FL setup, clients perform local training while the central server handles knowledge aggregation. Our proposed plug-in, DART, is integrated on the server side, positioned immediately after the aggregation step and before the global model is redistributed to clients.

\vspace{-0.1in} 

\paragraph{Client-Side Training.} Each client $c_k$ initializes its local model $f_{\w_k}$ using the global model, $f_{\w_0}$, provided by the server. It then performs $E$ epochs of local training on its private dataset $D_k$ following the standard protocol of the underlying FL algorithm. The updated model $f_{\w_k}$ is subsequently transmitted to the server. This process repeats in every communication round with the latest global model. \textbf{Unlike existing robust FL approaches, DART does not require clients to perform any robustness related computation.} Instead, clients adhere to their conventional training routines, while robustness enhancement is entirely offloaded to the server. \textbf{This allows DART enhanced systems to preserve the benefits of the underlying FL algorithm.} Other naive implementations aiming to reduce the resource cost of robust FL algorithms such as reducing the number of local epochs, or performing few robust iterations at the end of training, sacrifice accuracy, suffer from objective mismatch and fail to address the memory overhead.

\paragraph{Server-Side Aggregation and Robustness Enhancement.} After receiving the locally trained models from participating clients, the server performs aggregation following the protocol of the underlying federated learning system.

In standard FL, the aggregated global model $f_{\w_0}$ is directly transmitted back to clients for the next communication round. In contrast, when our DART plug-in is enabled, the server performs an additional robustness enhancement step after every $\robT$ global rounds. Specifically, the aggregated model undergoes robust training via the DART procedure, resulting in a robustness-enhanced model $f_{\DARTw}$. This process leverages a public, unlabeled dataset $D_0 \sim\outD$, which is independent of client data and therefore preserves privacy. \textbf{This significantly departs from traditional robust training, which requires access to the target data distribution.} The system level Algorithm is provided in the Appendix

\subsection{Data-Agnostic Robust Training}
\label{sec:DART}
We propose DART to preserve the privacy guarantees of FL by avoiding access to client data while enhancing robustness. Current approaches require the possession of data from the target learned data distribution $\sim \inD$, making them unsuitable for server-side deployment. In contrast, DART operates on a separate unlabeled dataset $D_0 \sim \outD$, and transforms the aggregated model $f_{\w_0}$ into a robust model $f_{\DARTw}$, while maintaining clean accuracy and improving robust accuracy. Even though DART operates on the server dataset $D_0 \sim \outD$, all performance evaluations, in terms of clean accuracy and robust accuracy, are conducted on samples drawn from $\inD$ and $\corrD$, respectively. \textbf{While DART is designed to enhance robustness in FL, it also applies to settings where original training data cannot be accessed.} Such situations may arise due to privacy constraints, prohibitive transmission costs associated with large datasets, or intellectual property restrictions.

\paragraph{DART Loss Function.} The DART loss is defined as a weighted sum of two components, $\dloss$ and $\closs$, each representing a learned objective:
\begin{equation}
    \DARTloss = \dloss + \alpha \closs
    \label{eq:DARTloss}
\end{equation}
where $\alpha$ is a weighting hyperparameter widely adopted in the robust training literature. The consistency loss $\closs$ promotes output similarity for semantically identical images under different perturbations. The distillation loss $\dloss$ ensures that the robust model $f_{\DARTw}$ retains the clean accuracy of the aggregated model $f_{\w_0}$, especially important since DART operates on a distributionally distinct dataset. Together, these losses guide the model to better predict corrupted inputs $\sim \corrD$ while preserving performance on clean data $\sim \inD$, resulting in an accurate and robust model with a favorable utility-robustness-efficiency tradeoff.

For DART training and validation, the public dataset $D_0$ is partitioned into disjoint subsets: $\DARTD$ for training and $\valD$ for validation. We apply early stopping based on the validation loss $\DARTloss$ computed on $\valD$. Next, we detail each loss component computed using the proxy dataset $D_0$.

\paragraph{Enhancing Consistency.} From each training input $\DARTx \in D_0 $, two augmented samples, $\augfirstx$ and $\augsecondx$, are generated using the data augmentation scheme described in the Appendix. These three images are passed through the student model to produce the predictions $\DARTp$, $\augfirstp$, and $\augsecondp$. We compute the consistency loss as follows:

\begin{equation}
    \closs = \operatorname{JS}\left(\DARTp, \augfirstp, \augsecondp\right)
\end{equation}
where $\operatorname{JS}$ denotes Jensen-Shannon divergence, a symmetric and smooth variant of the Kullback–Leibler ($\operatorname{KL}$) divergence. Minimizing $\closs$ by updating the model's parameters steers the model towards similar predictions for clean and augmented versions of the same input, aligning the model’s predictions for images representing the same information. \textbf{However, optimizing solely for $\closs$ compromises both utility and robustness, as it neglects to preserve the predictive capability acquired during client-side local training on the client’s data distribution.} To address this issue, DART incorporates a second loss, $\dloss$, to retain the model’s predictive knowledge.

\paragraph{Maintaining Accuracy.} 
\label{sec:maintaing}

We adopt a teacher-student framework, where the teacher model is fixed as the aggregated client pre-trained model, $f_{\w_\text{t}} = f_{\w_0}$, and the student model, $f_{\w_\text{s}}$, is updated from $\w_0$ to $\DARTw$. The consistency loss $\closs$ is evaluated using the student model. The input $\DARTx$ is passed through both teacher and student models to generate predictions, $\teacherp$ and $\studentp$. The distillation loss is then computed as:

\begin{equation}
    \dloss = \operatorname{KL}\left(\teacherp \parallel \studentp\right)
\end{equation}

Minimizing $\dloss$ encourages the student model to closely match the predictions of the fixed pre-trained teacher model. \textbf{This ensures that the student preserves the aggregated model's clean accuracy on the target data distribution.} The combination of preserving accuracy and enhancing consistency by minimizing $\DARTloss$ yields a robust model at zero additional client cost. \textbf{Despite being optimized on a distributionally mismatched proxy dataset, $\dloss$ remains effective in transferring the knowledge learned during local training, as the teacher’s soft predictions act as informative targets that guide the student model toward better generalization.} DART is illustrated in Fig.~\ref{fig:DART_DIAGRAM} and the algorithm is provided in the Appendix.

\subsection{Performance Bound}
Since DART operates on a proxy dataset following distribution $\outD$ to enhance robustness on the target dataset following $\inD$, we examine the effect of DART on clean accuracy, and equivalently clean risk. The proof is provided in the Appendix.

\begin{theorem}[Clean Student Risk Bound under Distillation]
\label{thm:main}
For any $\gamma \in (0, 1]$, the clean risk of the student model $\R_{\inD}(f_{\w_\text{s}})$ satisfies the following upper bound:
\begin{align}
    \R_{\inD}(f_{\w_\text{s}})
    &\le
    \R_{\inD}(f_{\w_\text{t}})
    + \theta_{\outD}(\gamma) \nonumber \\
    &\quad
    + \frac{2}{\gamma}\sqrt{\frac{\delta}{2}}
    + \frac{1}{2}\, d_{\mathcal{H}\Delta\mathcal{H}}(\inD, \outD)
    \label{eq:main}
\end{align}
\end{theorem}

\noindent where $\theta_{\outD}(\gamma) = \Pr_{x \sim \outD}[\,m_t(x) \le \gamma\,]$ is the margin tail of the teacher on proxy data $D_0$, where the margin is $m_t(x) = \max_{y_t^*}\teacherp(y_t^*|x) - \max_{y \neq y_t^*} \teacherp(y|x)$, $\delta = \mathbb{E}_{x \sim \outD}\!\left[\mathrm{KL}\!\left(\teacherp(\cdot|x) \,\Vert\, \studentp(\cdot|x)\right)\right]$ is the expected distillation divergence between teacher and student posteriors on $D_0 \sim \outD$ and $d_{\mathcal{H}\Delta\mathcal{H}}(\inD, \outD)$ is the $\mathcal{H}\!\Delta\!\mathcal{H}$-divergence~\cite{ben2010theory} between the target distribution $\inD$ and proxy distribution $\outD$.

\begin{table*}[t]
\centering
\setlength{\tabcolsep}{4pt}
\renewcommand{\arraystretch}{0.9}
\footnotesize
\caption{Impact of DART in enhancing the robustness of diverse standard FL methods. Clean accuracy is measured on CIFAR-10 ($\clnA$), and robust accuracy on CIFAR-10-C ($\robA^\mathrm{C}$) for ResNet-18. Accuracies on select CIFAR-10-C corruptions (contr, defoc, frost, g\_noise, glass, imp, motion, shot, zoom) are shown. 
DART significantly enhances the robustness (avg. $4.2\%$) of standard FL methods while incurring a small drop (avg. $1.8\%$) in clean accuracy at no  additional computational cost for the clients.}
\begin{tabular}{@{}l*{11}{c}@{}}
\toprule
Method & $\clnA (\%) $ & $\robA^\mathrm{C} (\%)$ & contr$(\%)$& defoc$(\%)$& frost$(\%)$& g\_noise$(\%)$& glass$(\%)$& imp$(\%)$& motion$(\%)$& shot$(\%)$& zoom$(\%)$\\
\midrule
\multicolumn{12}{c}{$\alpha_\text{IID} = 100$} \\
\midrule
FedAvg & \pmerr{90.3}{0.9} & \pmerr{70.5}{0.5} & \pmerr{68.2}{2.3} & \pmerr{75.8}{2.4} & \pmerr{72.2}{1.7} & \pmerr{47.8}{6.6} & \pmerr{46.0}{3.3} & \pmerr{55.6}{4.2} & \pmerr{71.8}{3.7} & \pmerr{58.3}{4.8} & \pmerr{71.6}{2.4}  \\

FedDyn & \pmerr{85.0}{1.0} & \pmerr{67.7}{1.3} & \pmerr{63.7}{2.4} & \pmerr{73.7}{2.9} & \pmerr{68.1}{2.7} & \pmerr{53.5}{6.2} & \pmerr{43.3}{6.9} & \pmerr{56.9}{2.6} & \pmerr{65.9}{4.2} & \pmerr{61.3}{5.0} & \pmerr{68.3}{3.4}  \\

FedNova & \pmerr{86.5}{0.5} & \pmerr{69.5}{0.8} & \pmerr{61.9}{0.9} & \pmerr{75.0}{0.4} & \pmerr{69.1}{1.3} & \pmerr{56.4}{2.0} & \pmerr{45.6}{2.4} & \pmerr{58.9}{0.9} & \pmerr{68.4}{0.9} & \pmerr{\textbf{63.1}}{1.6} & \pmerr{70.2}{0.9}  \\

FedProx & \pmerr{\textbf{91.4}}{0.3} & \pmerr{72.7}{1.0} & \pmerr{73.5}{1.6} & \pmerr{79.1}{1.0} & \pmerr{73.7}{1.3} & \pmerr{51.7}{2.8} & \pmerr{45.5}{2.5} & \pmerr{60.7}{2.1} & \pmerr{73.8}{1.1} & \pmerr{61.8}{2.5} & \pmerr{73.5}{1.5}  \\

\addlinespace[2pt]
\cmidrule(lr){1-12}
\addlinespace[2pt]

FedAvg+DART & \pmerr{89.2}{0.6} & \pmerr{77.4}{0.9} & \pmerr{73.8}{2.1} & \pmerr{84.8}{1.2} & \pmerr{75.2}{0.9} & \pmerr{64.0}{2.3} & \pmerr{57.7}{2.1} & \pmerr{69.2}{1.8} & \pmerr{80.1}{2.3} & \pmerr{71.7}{1.5} & \pmerr{82.5}{1.4}  \\

FedDyn+DART & \pmerr{83.2}{1.8} & \pmerr{71.2}{1.2} & \pmerr{67.4}{1.9} & \pmerr{78.6}{1.6} & \pmerr{67.9}{1.4} & \pmerr{59.2}{3.0} & \pmerr{51.5}{4.7} & \pmerr{65.2}{1.4} & \pmerr{72.8}{2.2} & \pmerr{66.2}{2.0} & \pmerr{75.7}{0.7}  \\

FedNova+DART & \pmerr{85.1}{0.6} & \pmerr{72.6}{0.8} & \pmerr{66.7}{1.4} & \pmerr{81.0}{0.8} & \pmerr{68.1}{1.1} & \pmerr{60.0}{1.0} & \pmerr{51.0}{1.6} & \pmerr{64.5}{1.0} & \pmerr{74.3}{1.4} & \pmerr{67.0}{0.8} & \pmerr{78.1}{0.9}  \\

FedProx+DART & \pmerr{90.2}{0.4} & \pmerr{\textbf{79.0}}{0.8} & \pmerr{\textbf{77.7}}{2.0} & \pmerr{\textbf{86.1}}{0.7} & \pmerr{\textbf{77.3}}{1.4} & \pmerr{\textbf{65.6}}{2.7} & \pmerr{\textbf{57.9}}{3.6} & \pmerr{\textbf{72.6}}{1.9} & \pmerr{\textbf{81.8}}{1.0} & \pmerr{\textbf{73.7}}{1.9} & \pmerr{\textbf{83.5}}{0.8}  \\

\midrule
\multicolumn{12}{c}{$\alpha_\text{IID} = 10$} \\
\midrule

FedAvg & \pmerr{89.5}{0.9} & \pmerr{69.6}{1.8} & \pmerr{68.6}{1.7} & \pmerr{76.0}{1.0} & \pmerr{69.7}{2.5} & \pmerr{46.6}{7.4} & \pmerr{43.6}{3.2} & \pmerr{54.3}{5.9} & \pmerr{72.6}{1.1} & \pmerr{56.6}{6.3} & \pmerr{71.1}{2.2}  \\

FedDyn & \pmerr{83.9}{1.7} & \pmerr{66.7}{1.1} & \pmerr{60.9}{4.0} & \pmerr{72.6}{1.8} & \pmerr{65.9}{1.9} & \pmerr{52.5}{6.6} & \pmerr{43.5}{6.8} & \pmerr{57.2}{2.4} & \pmerr{66.6}{0.6} & \pmerr{59.7}{4.9} & \pmerr{67.1}{1.7}  \\

FedNova & \pmerr{85.7}{0.3} & \pmerr{67.9}{0.5} & \pmerr{60.1}{1.3} & \pmerr{73.6}{0.9} & \pmerr{67.1}{0.5} & \pmerr{54.2}{1.6} & \pmerr{44.6}{1.0} & \pmerr{57.3}{1.8} & \pmerr{65.6}{0.7} & \pmerr{61.1}{1.6} & \pmerr{68.4}{1.1}  \\

FedProx & \pmerr{\textbf{90.6}}{0.4} & \pmerr{71.9}{1.0} & \pmerr{71.9}{2.1} & \pmerr{79.4}{1.0} & \pmerr{71.7}{0.9} & \pmerr{50.0}{4.2} & \pmerr{45.6}{3.1} & \pmerr{59.1}{2.4} & \pmerr{73.9}{1.1} & \pmerr{59.7}{3.1} & \pmerr{74.9}{1.5}  \\

\addlinespace[2pt]
\cmidrule(lr){1-12}
\addlinespace[2pt]

FedAvg+DART & \pmerr{88.3}{0.7} & \pmerr{75.8}{1.5} & \pmerr{71.8}{1.0} & \pmerr{83.4}{1.2} & \pmerr{73.2}{2.6} & \pmerr{62.1}{3.9} & \pmerr{54.7}{4.1} & \pmerr{67.4}{2.4} & \pmerr{78.9}{1.7} & \pmerr{70.0}{3.1} & \pmerr{81.1}{1.3}  \\

FedDyn+DART & \pmerr{82.7}{1.4} & \pmerr{69.5}{2.4} & \pmerr{64.1}{3.1} & \pmerr{77.9}{1.8} & \pmerr{64.7}{3.8} & \pmerr{56.8}{4.4} & \pmerr{47.8}{4.1} & \pmerr{63.0}{3.5} & \pmerr{71.4}{3.2} & \pmerr{63.7}{3.9} & \pmerr{74.6}{1.8}  \\

FedNova+DART & \pmerr{84.7}{0.3} & \pmerr{71.8}{0.4} & \pmerr{66.7}{0.5} & \pmerr{80.4}{0.5} & \pmerr{67.3}{0.7} & \pmerr{58.2}{1.6} & \pmerr{48.9}{1.4} & \pmerr{64.0}{0.4} & \pmerr{74.2}{0.6} & \pmerr{65.5}{1.2} & \pmerr{77.2}{0.6}  \\

FedProx+DART & \pmerr{88.7}{0.6} & \pmerr{\textbf{77.2}}{2.0} & \pmerr{\textbf{74.7}}{3.7} & \pmerr{\textbf{84.6}}{0.8} & \pmerr{\textbf{74.5}}{3.4} & \pmerr{\textbf{64.5}}{4.3} & \pmerr{\textbf{56.8}}{4.3} & \pmerr{\textbf{70.7}}{2.5} & \pmerr{\textbf{80.0}}{2.0} & \pmerr{\textbf{72.0}}{3.7} & \pmerr{\textbf{82.1}}{1.2}  \\

\midrule
\multicolumn{12}{c}{$\alpha_\text{IID} = 1$} \\
\midrule

FedAvg & \pmerr{87.8}{1.4} & \pmerr{65.6}{1.9} & \pmerr{63.8}{1.8} & \pmerr{73.0}{1.4} & \pmerr{65.4}{4.0} & \pmerr{42.1}{4.8} & \pmerr{39.2}{4.3} & \pmerr{50.1}{3.1} & \pmerr{65.4}{2.7} & \pmerr{52.3}{5.0} & \pmerr{66.4}{2.5}  \\

FedDyn & \pmerr{80.7}{1.0} & \pmerr{64.3}{1.1} & \pmerr{58.9}{2.0} & \pmerr{69.9}{2.3} & \pmerr{62.6}{2.8} & \pmerr{52.7}{4.1} & \pmerr{44.1}{5.6} & \pmerr{53.9}{2.5} & \pmerr{62.6}{3.9} & \pmerr{58.8}{3.3} & \pmerr{65.2}{3.6}  \\

FedNova & \pmerr{82.7}{0.6} & \pmerr{64.9}{0.8} & \pmerr{55.3}{0.8} & \pmerr{69.3}{0.9} & \pmerr{64.3}{0.4} & \pmerr{54.1}{1.6} & \pmerr{45.6}{1.7} & \pmerr{55.3}{1.3} & \pmerr{60.1}{1.3} & \pmerr{60.2}{1.4} & \pmerr{63.4}{0.7}  \\

FedProx & \pmerr{\textbf{88.9}}{1.1} & \pmerr{70.2}{2.0} & \pmerr{68.1}{1.5} & \pmerr{77.1}{2.1} & \pmerr{69.4}{2.9} & \pmerr{51.7}{5.9} & \pmerr{42.9}{5.7} & \pmerr{58.7}{5.0} & \pmerr{70.0}{3.4} & \pmerr{60.8}{4.7} & \pmerr{72.2}{2.3}  \\

\addlinespace[2pt]
\cmidrule(lr){1-12}
\addlinespace[2pt]

FedAvg+DART & \pmerr{83.2}{3.2} & \pmerr{68.4}{5.5} & \pmerr{63.9}{5.7} & \pmerr{76.1}{4.4} & \pmerr{64.8}{5.4} & \pmerr{56.2}{7.0} & \pmerr{46.6}{9.2} & \pmerr{60.8}{6.8} & \pmerr{67.8}{6.3} & \pmerr{63.0}{6.9} & \pmerr{71.8}{5.3}  \\

FedDyn+DART & \pmerr{79.1}{1.5} & \pmerr{67.2}{1.6} & \pmerr{63.1}{3.3} & \pmerr{74.4}{2.5} & \pmerr{63.5}{3.2} & \pmerr{56.2}{3.7} & \pmerr{48.8}{3.6} & \pmerr{59.5}{0.9} & \pmerr{67.9}{3.0} & \pmerr{62.2}{3.1} & \pmerr{72.4}{2.5}  \\

FedNova+DART & \pmerr{80.3}{0.7} & \pmerr{67.6}{1.1} & \pmerr{59.4}{1.9} & \pmerr{74.6}{1.1} & \pmerr{63.6}{1.6} & \pmerr{58.2}{1.5} & \pmerr{\textbf{50.1}}{1.6} & \pmerr{60.3}{1.6} & \pmerr{67.1}{1.8} & \pmerr{63.8}{1.2} & \pmerr{71.3}{1.9}  \\

FedProx+DART & \pmerr{87.1}{0.9} & \pmerr{\textbf{73.9}}{1.5} & \pmerr{\textbf{71.0}}{1.8} & \pmerr{\textbf{82.0}}{1.2} & \pmerr{\textbf{70.6}}{1.7} & \pmerr{\textbf{60.8}}{3.3} & \pmerr{49.7}{2.4} & \pmerr{\textbf{67.4}}{2.0} & \pmerr{\textbf{74.8}}{2.4} & \pmerr{\textbf{68.5}}{2.8} & \pmerr{\textbf{78.6}}{1.3}  \\

\bottomrule
\end{tabular}

\label{tab:plugin_result}
\end{table*}

\paragraph{Interpretation.} A tighter upper bound on student clean risk can be achieved by minimizing the expected KL divergence $\delta$, choosing a server dataset distribution similar to the client distribution such that $\mathcal{H}\!\Delta\!\mathcal{H}$-divergence is small and only considering server samples with high output probability margin. In practice, $R_{\inD}(f_{\w_\text{s}})$ approaches $R_{\inD}(f_{\w_\text{t}})$ due to student initialization and the choice of an appropriately small learning rate.

%% file: sec/6_results.tex
\section{Results}
\label{sec:Results}

\subsection{Setup}
\label{sec:setup}

We adopt the following setup for our experiments. Deviations are documented in the relevant subsections and clarified in figure and table captions where applicable.

\textbf{Datasets and Models:} We use CIFAR-10~\cite{krizhevsky2009learning}~$\sim \inD$ as the private client dataset, while CIFAR-100~\cite{krizhevsky2009learning} or Tiny ImageNet~\cite{le2015tiny} serves as the server dataset $D_0 \sim \outD$. To model data heterogeneity, client data are partitioned using a Dirichlet distribution with parameter $\alpha_\text{IID}$, where smaller values indicate greater heterogeneity. Additional results using CIFAR-100 as the client dataset are reported in the Appendix. Our dataset setup follows prior work~\cite{li2021model, zhang2022federated, yang2024fedas, xie2024perada}. For corruption robustness, we evaluate on CIFAR-10-C~\cite{hendrycks2019benchmarking}, CIFAR-10-$\bar{\mathrm{C}}$~\cite{mintun2021interaction}, and CIFAR-10-P~\cite{hendrycks2019benchmarking}. All clients use ResNet-18~\cite{he2016deep}, consistent with prior FL studies~\cite{gao2024device, seo2024relaxed}. To assess generalization, we additionally evaluate MobileNet~\cite{howard2017mobilenets} and VGG-16~\cite{simonyan2014very}, and include experiments with synthetic server data in the Appendix.

\textbf{Metrics:} Clean accuracy evaluates performance on clean data~$\sim \inD$, while robust accuracy evaluates performance on corrupted data~$\sim \corrD$. The average per-client training time $\trainTime$, energy $\trainEnergy$, and memory $\trainMemory$ are the efficiency metrics of interest. 
The resource model and experiments accounting for server-side costs can be found in the Appendix.

\begin{table*}[t]
\centering
\setlength{\tabcolsep}{4pt}      
\renewcommand{\arraystretch}{0.9}
\footnotesize                   
\caption{Impact of DART on robustness across diverse benchmarks. Robust accuracy is evaluated using CIFAR-10-$\bar{\mathrm{C}}$ ($\robA^{\bar{\mathrm{C}}}$) and the flip probability ($p_\mathrm{flip}$) is evaluated on CIFAR-10-P for ResNet-18.  DART significantly improves $\robA^{\bar{\mathrm{C}}}$ (up to $3.7\%$) and reduces $p_\mathrm{flip}$ (up to $0.8\%$) relative to standard FL methods. All improvements are achieved with zero additional client cost.}
\label{tab:plugin_results_bar}

\begin{tabular*}{\textwidth}{@{\extracolsep{\fill}} lcccccccc @{}}
\toprule
\multirow{2}{*}{Method}
  & \multicolumn{2}{c}{$\alpha_\text{IID}=100$}
  & \multicolumn{2}{c}{$\alpha_\text{IID}=10$}
  & \multicolumn{2}{c}{$\alpha_\text{IID}=1$}
  & \multicolumn{2}{c}{$\alpha_\text{IID}=0.1$} \\
\cmidrule(lr){2-3}\cmidrule(lr){4-5}\cmidrule(lr){6-7}\cmidrule(lr){8-9}
 & $\robA^{\bar{\mathrm{C}}}(\%)$ & $p_\mathrm{flip}$$(\%)$ & $\robA^{\bar{\mathrm{C}}}(\%)$ & $p_\mathrm{flip}$$(\%)$ & $\robA^{\bar{\mathrm{C}}}(\%)$ & $p_\mathrm{flip}$$(\%)$ & $\robA^{\bar{\mathrm{C}}}(\%)$ & $p_\mathrm{flip}$$(\%)$ \\
\midrule
FedAvg        & 69.0 $\pm$ 1.0 & 4.9 & 68.3 $\pm$ 0.7 & 5.0 & 64.0 $\pm$ 1.2 & 5.0 & 36.0 $\pm$ 11.8 & 6.0 \\
FedProx       & 71.1 $\pm$ 0.8 & 4.4 & 70.6 $\pm$ 0.5 & 4.2 & 67.6 $\pm$ 1.8 & 4.2 & 53.2 $\pm$ 1.3 & 4.8 \\
FedDyn       & 65.2 $\pm$ 0.5 & 5.0 & 64.5 $\pm$ 1.1 & 4.9 & 62.3 $\pm$ 1.1 & 5.3 & 31.0 $\pm$  6.3 & 3.0 \\
FedNova       & 67.8 $\pm$ 0.4 & 4.6 & 66.8 $\pm$ 0.5 & 4.7 & 63.9 $\pm$ 0.7 & 4.8 & 43.8 $\pm$ 3.2 & 4.8 \\
\addlinespace[2pt]
\cmidrule(lr){1-9}
\addlinespace[2pt]
FedAvg{+}DART   & 72.8 $\pm$ 0.8 & {4.1}
                & 71.9 $\pm$ 1.1 & 4.2
                & 63.5 $\pm$ 4.7 & 4.3
                & 31.2 $\pm$ 14.1            & 5.3 \\
FedProx{+}DART  & \textbf{74.0} $\pm$ 1.0 & \textbf{3.8}
                & \textbf{72.8} $\pm$ 0.9 & \textbf{3.7}
                & \textbf{69.3} $\pm$ 0.8 & \textbf{3.7}
                & \textbf{53.5} $\pm$ 4.0 & \textbf{4.2} \\
FedDyn{+}DART   & 66.8 $\pm$ 1.4 & {4.7}
                & 66.0 $\pm$ 1.8 & 4.6
                & 63.4 $\pm$ 2.1 & 4.8
                & 21.3 $\pm$ 13.1            & 3.1 \\
FedNova{+}DART   & 68.9 $\pm$ 0.5 & {4.3}
                & 68.2 $\pm$ 0.3 & 4.4
                & 64.0 $\pm$ 0.9 & 4.7
                & 38.5 $\pm$ 5.1            & 4.6 \\
\bottomrule
\end{tabular*}

\end{table*}

\begin{table*}[t]
\centering
\setlength{\tabcolsep}{0pt}
\renewcommand{\arraystretch}{0.9}
\footnotesize
\caption{CIFAR-10 clean accuracy $\clnA$, CIFAR-10-C robust accuracy $\robA^{{\mathrm{C}}}$ and CIFAR-10-$\bar{\mathrm{C}}$ robust accuracy $\robA^{\bar{\mathrm{C}}}$ for robust FL methods and DART-enhanced FL. \textbf{Bold} and \underline{underline} denote the best and second-best results, respectively. DART-enhanced methods deliver robust solutions while requiring only a fraction of the time, energy, and memory.}
\label{tab:robust_efficiency}

\begin{tabular*}{\textwidth}{@{\extracolsep{\fill}} l ccc ccc ccc @{}}
\toprule
\multirow{2}{*}{Method}
  & \multicolumn{3}{c}{$\alpha_{\text{IID}}=100$}
  & \multicolumn{3}{c}{$\alpha_{\text{IID}}=10$}
  & \multicolumn{3}{c}{Resources} \\
\cmidrule(lr){2-4}\cmidrule(lr){5-7}\cmidrule(lr){8-10}
& $\clnA(\%)$ & $\robA^{{\mathrm{C}}}(\%)$ & $\robA^{\bar{\mathrm{C}}}(\%)$
& $\clnA(\%)$ & $\robA^{{\mathrm{C}}}(\%)$ & $\robA^{\bar{\mathrm{C}}}(\%)$
& $\trainTime$ & $\trainEnergy$ & $\trainMemory$ \\
\midrule
FedAFA
  & $89.9 \pm 0.9$ & $80.6 \pm 0.4$ & $75.5 \pm 0.4$
  & $88.4 \pm 0.9$ & $79.6 \pm 2.3$ & $74.3 \pm 2.0$
  & $\times 1.9$ & $\times 1.9$ & $\times 1.7$ \\
FedAugMix
  & $89.3 \pm 1.7$ & $\underline{82.5} \pm 1.7$ & $\underline{76.9} \pm 1.9$
  & $\mathbf{89.1} \pm 0.8$ & $\mathbf{82.5} \pm 1.5$ & $\mathbf{77.0} \pm 1.1$
  & $\times 2.6$ & $\times 2.9$ & $\times 2.6$ \\
FedPrime
  & $86.6 \pm 0.8$ & $\mathbf{82.9} \pm 0.8$ & $\mathbf{76.3} \pm 0.6$
  & $85.0 \pm 1.9$ & $\underline{81.0} \pm 2.0$ & $\underline{74.6} \pm 2.1$
  & $\times 21.5$ & $\times 5.7$ & $\times 2.6$ \\
\addlinespace[2pt]
\cmidrule(lr){1-10}
\addlinespace[2pt]
FedAvg{+}DART
  & $\underline{89.2} \pm 0.6$ & $77.4 \pm 0.9$ & $72.8 \pm 0.8$
  & $88.3 \pm 0.7$ & $75.8 \pm 1.5$ & $71.8 \pm 1.1$
  & $\mathbf{\times 1.0}$ & $\mathbf{\times 1.0}$ & $\mathbf{\times 1.0}$ \\
FedProx{+}DART
  & $\mathbf{90.2} \pm 0.4$ & $79.0 \pm 0.8$ & $74.0 \pm 1.0$
  & $\underline{88.7} \pm 0.6$ & $77.2 \pm 2.0$ & $72.8 \pm 0.9$
  & $\underline{\times 1.1}$ & $\underline{\times 1.1}$ & $\underline{\times 1.3}$ \\
\end{tabular*}

\vspace{0.6em}

\begin{tabular*}{\textwidth}{@{\extracolsep{\fill}} l ccc ccc ccc @{}}
\toprule
\multirow{2}{*}{Method}
  & \multicolumn{3}{c}{$\alpha_{\text{IID}}=1$}
  & \multicolumn{3}{c}{$\alpha_{\text{IID}}=0.1$}
  & \multicolumn{3}{c}{Resources} \\
\cmidrule(lr){2-4}\cmidrule(lr){5-7}\cmidrule(lr){8-10}
& $\clnA(\%)$ & $\robA^{{\mathrm{C}}}(\%)$ & $\robA^{\bar{\mathrm{C}}}(\%)$
& $\clnA(\%)$ & $\robA^{{\mathrm{C}}}(\%)$ & $\robA^{\bar{\mathrm{C}}}(\%)$
& $\trainTime$ & $\trainEnergy$ & $\trainMemory$ \\
\midrule
FedAFA
  & $\underline{86.7} \pm 1.4$ & $\underline{75.6} \pm 2.6$ & $70.1 \pm 2.2$
  & $\underline{48.3} \pm 13.0$ & $\underline{41.5} \pm 12.8$ & $\underline{37.8} \pm 12.4$
  & $\times 1.9$ & $\times 1.9$ & $\times 1.7$ \\
FedAugMix
  & $84.6 \pm 2.6$ & $\mathbf{76.6} \pm 3.1$ & $\underline{71.0} \pm 3.0$
  & $44.2 \pm 14.8$ & $37.5 \pm 13.0$ & $34.3 \pm 12.7$
  & $\times 2.6$ & $\times 2.9$ & $\times 2.6$ \\
FedPrime
  & $79.6 \pm 2.2$ & $74.5 \pm 2.7$ & $67.3 \pm 3.4$
  & $30.3 \pm 13.2$ & $28.0 \pm 12.6$ & $25.4 \pm 11.6$
  & $\times 21.5$ & $\times 5.7$ & $\times 2.6$ \\
\addlinespace[2pt]
\cmidrule(lr){1-10}
\addlinespace[2pt]
FedAvg{+}DART
  & $83.2 \pm 3.2$ & $68.4 \pm 5.5$ & $63.5 \pm 4.7$
  & $43.0 \pm 15.8$ & $32.9 \pm 14.7$ & $31.3 \pm 14.1$
  & $\mathbf{\times 1.0}$ & $\mathbf{\times 1.0}$ & $\mathbf{\times 1.0}$ \\
FedProx{+}DART
  & $\mathbf{87.1} \pm 1.0$ & $73.9 \pm 1.5$ & $\mathbf{69.3} \pm 0.8$
  & $\mathbf{66.9} \pm 4.3$ & $\mathbf{58.7} \pm 2.1$ & $\mathbf{53.5} \pm 4.0$
  & $\underline{\times 1.1}$ & $\underline{\times 1.1}$ & $\underline{\times 1.3}$ \\
\bottomrule
\end{tabular*}
\end{table*}

\textbf{Implementation Details:} Results are shown for a 100-client FL setup with a 0.1 client participation rate, following prior work~\cite{zhang2023fedcr, zhang2024improving}. The loss weighting parameter $\alpha$ is set to 10 to balance $\dloss$ and $\closs$ (selection details in in the Appendix. We use a learning rate of 0.1 on the client side and 0.001 for DART on the server. The number of local epochs per round is $E=1$, the total number of global rounds is $G=1000$, the DART update period is $\robT=200$, the maximum number of DART epochs is $\maxT=200$, and early stopping patience is $\valT=3$.

\textbf{Baselines:} We augment popular FL methods such as FedAvg~\cite{mcmahan2017communication}, FedDyn~\cite{acar2021federated}, FedNova~\cite{wang2020tackling}, and  FedProx~\cite{li2020federated} with DART. We also compare to the common corruption robust FL method FedAugMix~\cite{fang2023robust}, and to FL extensions of robust training methods (PRIME~\cite{modas2022prime} and AFA~\cite{vaish2024fourier}).

\textbf{Hardware:} Each client is represented by a NVIDIA Jetson Orin Nano~\cite{nvidiaJetsonNano} which has an Ampere GPU, a 6-core ARM CPU, 8GB RAM, and delivers up to 67 TOPS. 

\textbf{Software:} Power is measured using Jetson Stats~\cite{jetson-stats}, and energy is computed by integrating power over time, approximated as the product of sampled power and time intervals.

\subsection{DART-enhanced FL vs. Clean FL}
\label{sec:DARTvsClean}

\begin{figure*}[h]
    \centering
    \begin{subfigure}{0.24\linewidth}
        \includegraphics[width=\linewidth]{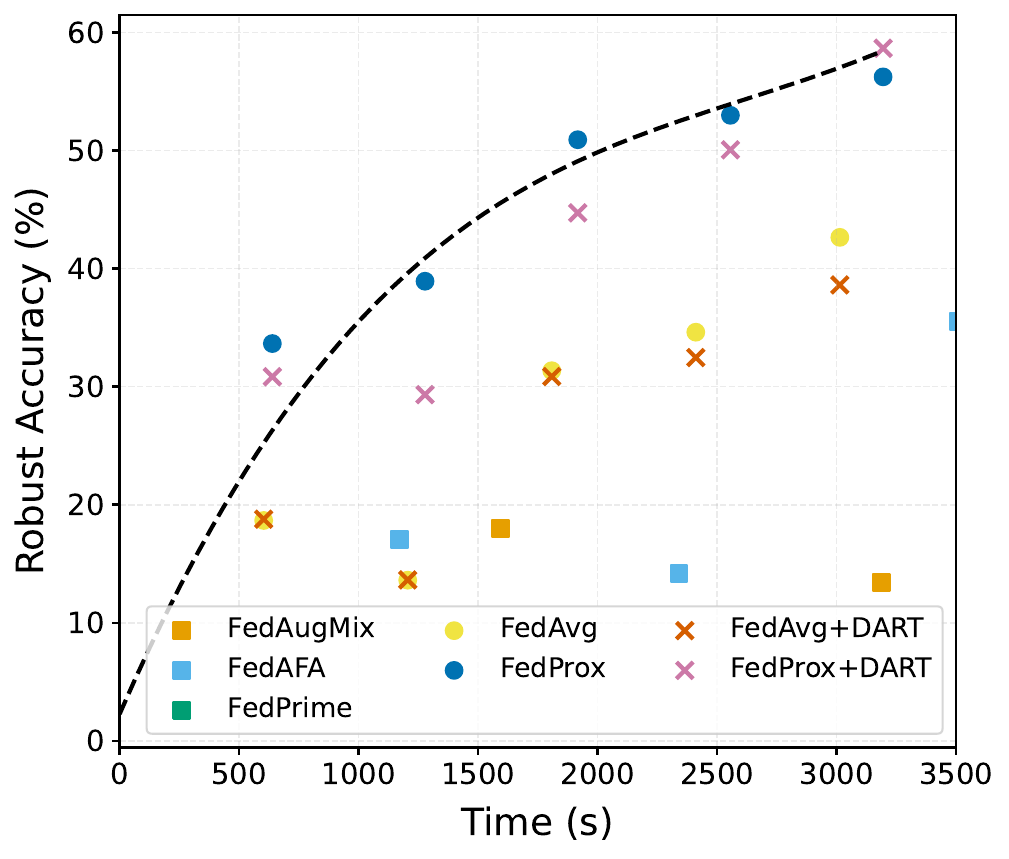}
        \caption{$\alpha_\text{IID} = 0.1$}
    \end{subfigure}
    \hfill
    \begin{subfigure}{0.24\linewidth}
        \includegraphics[width=\linewidth]{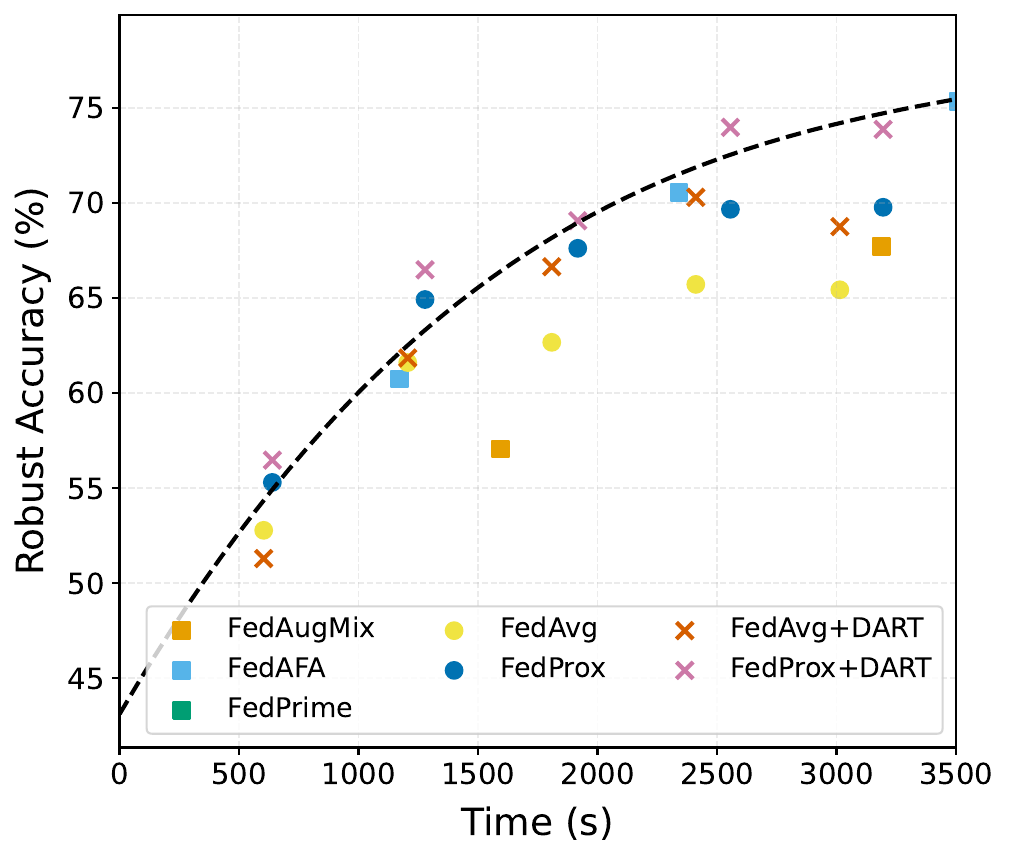}
        \caption{$\alpha_\text{IID} = 1$}
    \end{subfigure}
    \hfill
    \begin{subfigure}{0.24\linewidth}
        \includegraphics[width=\linewidth]{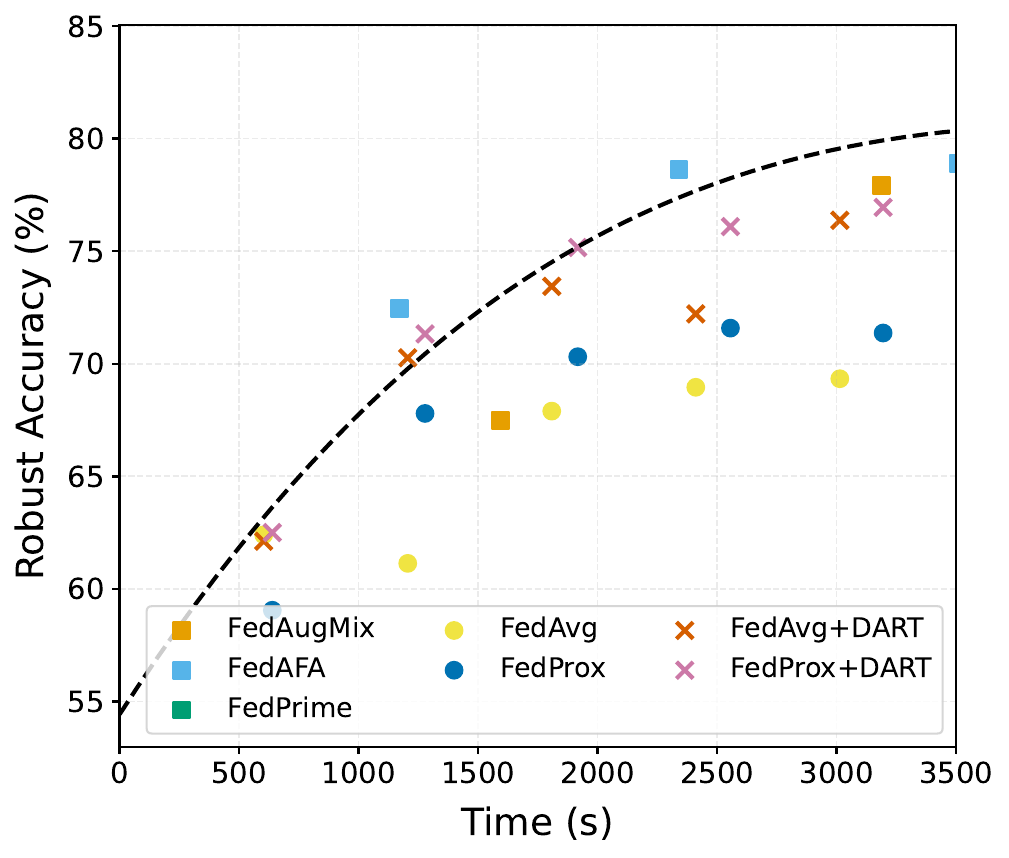}
        \caption{$\alpha_\text{IID} = 10$}
    \end{subfigure}
    \hfill
    \begin{subfigure}{0.24\linewidth}
        \includegraphics[width=\linewidth]{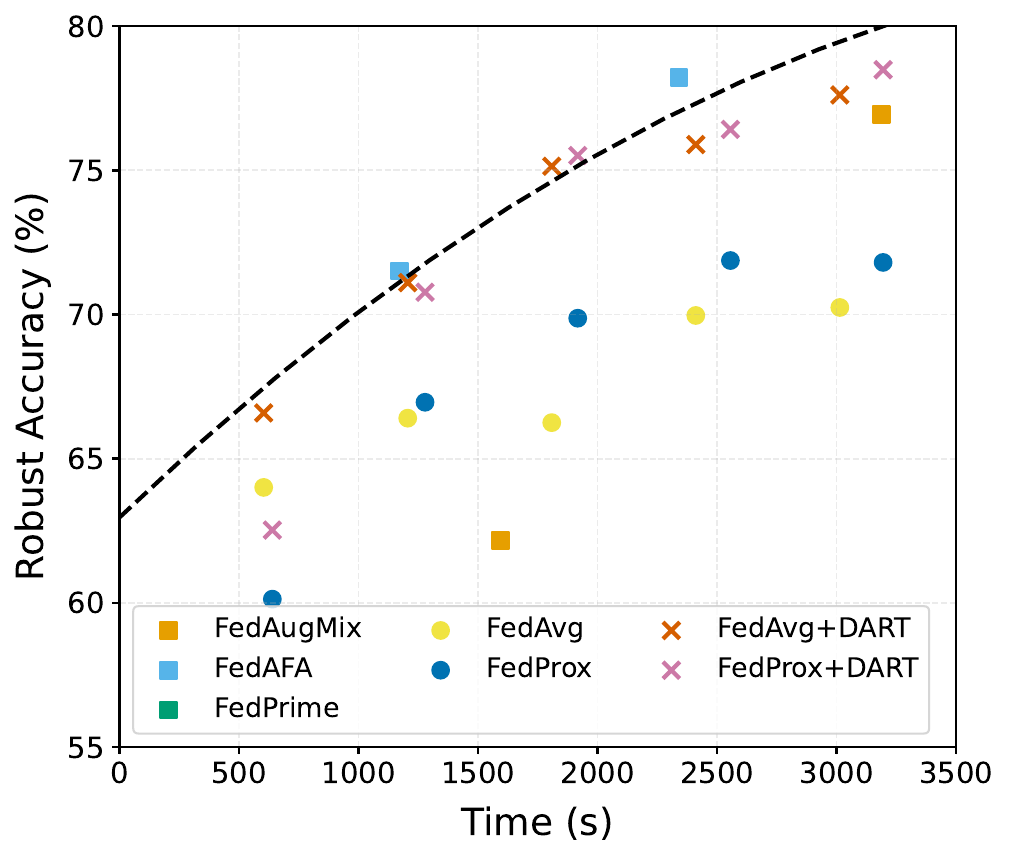}
        \caption{$\alpha_\text{IID} = 100$}
    \end{subfigure}
    \hfill
    \centering
    \begin{subfigure}{0.24\linewidth}
        \includegraphics[width=\linewidth]{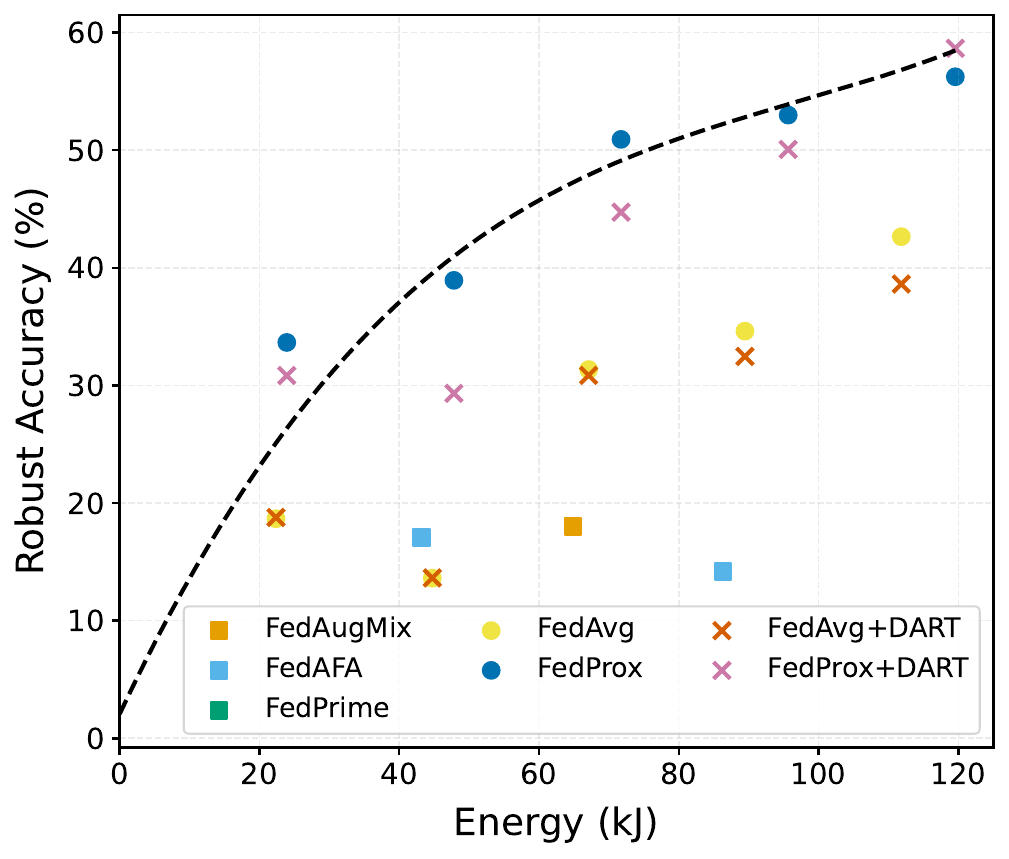}
        \caption{$\alpha_\text{IID} = 0.1$}
    \end{subfigure}
    \hfill
    \begin{subfigure}{0.24\linewidth}
        \includegraphics[width=\linewidth]{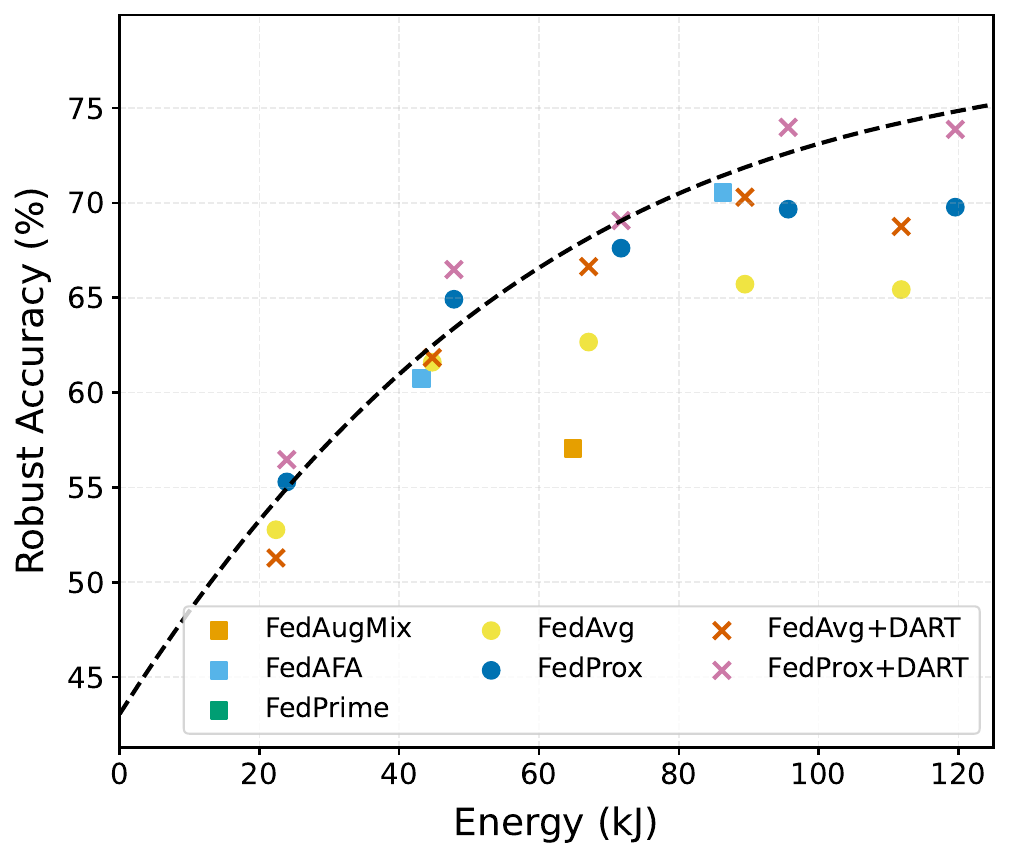}
        \caption{$\alpha_\text{IID} = 1$}
    \end{subfigure}
    \hfill
    \begin{subfigure}{0.24\linewidth}
        \includegraphics[width=\linewidth]{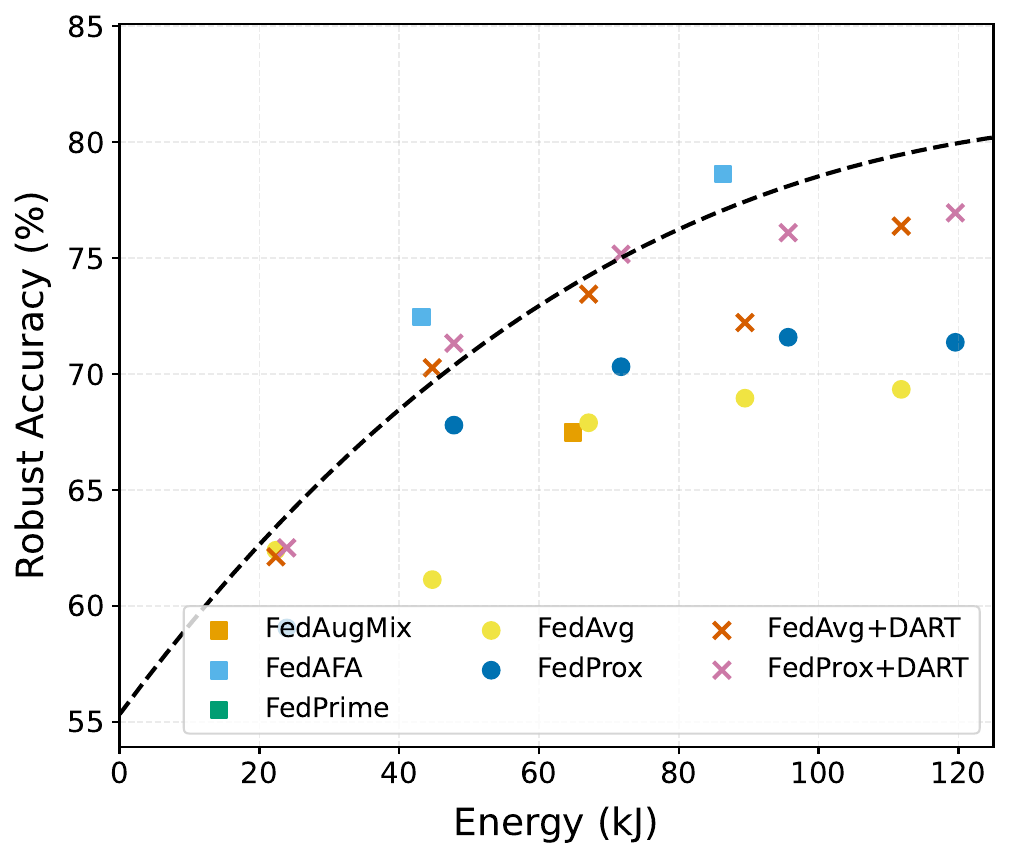}
        \caption{$\alpha_\text{IID} = 10$}
    \end{subfigure}
    \hfill
    \begin{subfigure}{0.24\linewidth}
        \includegraphics[width=\linewidth]{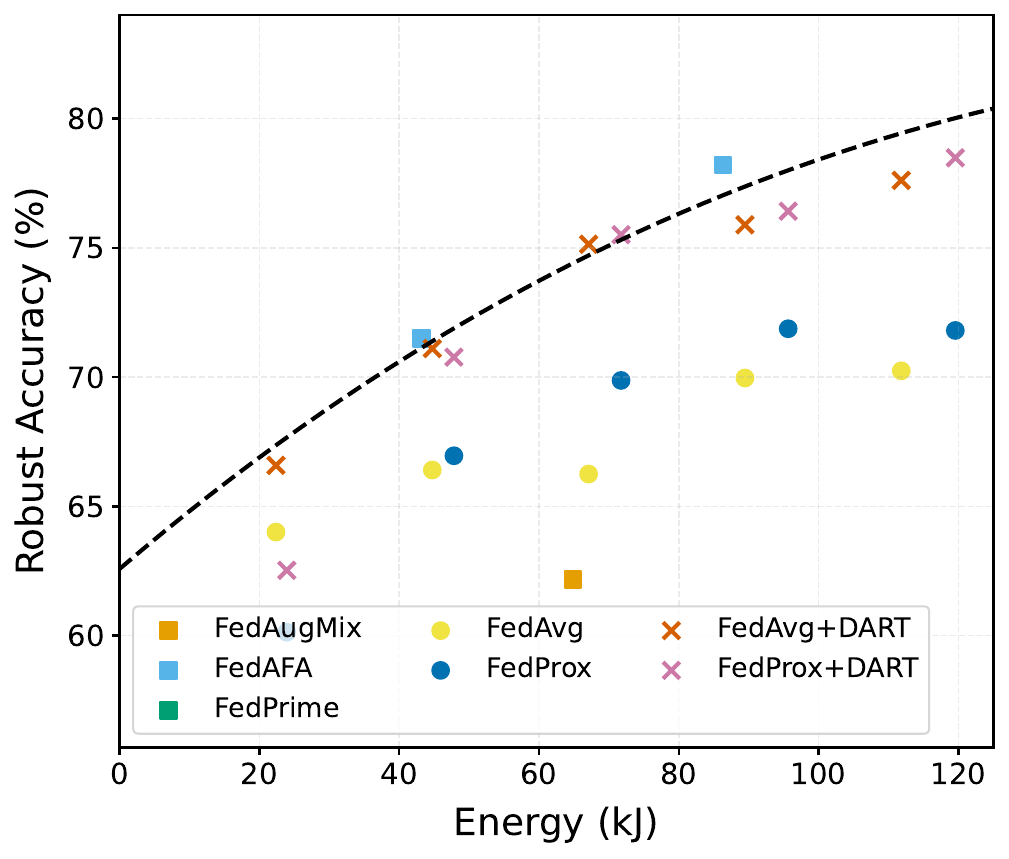}
        \caption{$\alpha_\text{IID} = 100$}
    \end{subfigure}
    \caption{ResNet-18 CIFAR-10-C robust accuracy as a function of training time and energy. Across all data heterogeneity $\alpha_\text{IID}$ levels, DART-enhanced methods consistently lie near the Pareto frontier (dashed curve), demonstrating a favorable robustness-efficiency tradeoff.}
    \label{fig:pareto}
\vskip -0.2in
\end{figure*}

We examine the impact of incorporating DART on the server side when performing FL with popular FL methods. Table~\ref{tab:plugin_result} shows that augmenting these methods with DART significantly improves performance on the CIFAR-10-C benchmark. Across varying heterogeneity levels $\alpha_\text{IID}$, DART improves robustness by an average of $4.2\%$ and up to $6.9\%$, with only a $1.8\%$ average reduction in clean accuracy, yielding a favorable utility–robustness trade-off at zero additional client cost. DART improves robustness across most CIFAR-10-C corruption types. Full corruption results are provided in the Appendix. Additionally, FedProx performs well under increased data heterogeneity, and this advantage is preserved after integrating DART, indicating that DART complements existing FL methods.

Table~\ref{tab:plugin_results_bar} reports robustness results on CIFAR-10-$\bar{\mathrm{C}}$ and CIFAR-10-P, two widely used corruption-robustness benchmarks. Incorporating DART on the server side consistently improves robustness, yielding an average gain of $1.8\%$ for FedProx across data heterogeneity levels, and up to $3.7\%$ for FedAvg on CIFAR-10-$\bar{\mathrm{C}}$. Moreover, models enhanced with DART exhibit, on average, a $0.4\%$ lower flipping probability, and up to $0.8\%$. Together, these results demonstrate that DART strengthens the robustness of FL methods while imposing zero additional client overhead.

\subsection{DART-enhanced FL vs. Robust FL}
We compare DART-enhanced FL with existing corruption robust FL methods, showing that DART achieves substantial resource savings while incurring a minor performance drop. DART-enhanced methods produce solutions that balance utility, robustness and client-side efficiency.

Table~\ref{tab:robust_efficiency} summarizes the utility, robustness and efficiency comparison between DART-enhanced and robust FL methods. DART achieves comparable or superior robustness while requiring significantly fewer computational resources. Specifically, DART-based methods reduce time, energy, and memory consumption by $1.9\text{–}21.5\times$, $1.9\text{–}5.7\times$, and $1.7\text{–}2.6\times$, respectively. Under low non-IID heterogeneity, robust FL baselines outperform DART-enhanced methods by only $3\%$ on average in robustness. At higher heterogeneity levels, DART-enhanced FedProx surpasses the robustness of all robust FL baselines. In terms of clean accuracy, DART-enhanced approaches generally outperform robust FL methods while maintaining lower resource costs. Overall, these results highlight DART’s favorable utility–robustness–efficiency trade-off, making it highly suitable across diverse heterogeneity regimes.

Fig.~\ref{fig:pareto} illustrate robustness as a function of time and energy, with a polynomial fit to the Pareto frontier highlighting optimal trade-offs. Across varying resource budgets and data heterogeneity levels, DART-enhanced methods consistently lie on or near the Pareto frontier. This indicates that, regardless of resource budget or heterogeneity level, DART-enhanced approaches deliver optimal or near-optimal robustness–efficiency trade-offs, making them highly practical for real-world deployment. In contrast, robust FL methods such as FedAFA approach optimality only under low heterogeneity. This result shows that simply reducing global rounds to save time and energy in robust FL methods fails to achieve optimal robust solutions under varying heterogeneity, whereas DART-enhanced FL remains close to the Pareto frontier under the same budget.

\begin{table}[h]
\centering
\caption{FedAvg ResNet-18 accuracy on CIFAR-10 and CIFAR-10-C for varying server-dataset ($D_0$) distributions and sizes. DART’s performance remains consistent across both distribution
shifts and dataset-size variations.}
\label{table:server_dataset_results}
\resizebox{0.85\linewidth}{!}{%
\begin{tabular}{lcccc}
\toprule
Server Dataset & \# of images & Clean & Robust & Average \\
\midrule

\multirow{2}{*}{CIFAR-100} 
  & 25k  & 91.72 & 80.12 & 85.92 \\
  & 50k  & \textbf{91.98} & \textbf{80.72} & \textbf{86.35} \\ \cmidrule(lr){1-5}

\multirow{4}{*}{Tiny ImageNet
} 
  & 25k  & 91.97 & 80.38 & 86.18 \\
  & 50k  & 92.03 & 80.92 & 86.48 \\
  & 75k  & \textbf{92.53} & \textbf{81.82} & \textbf{87.18} \\
  & 100k & 92.48 & 81.64 & 87.06 \\
\bottomrule
\end{tabular}
}
\end{table}

\subsection{Impact of Server Dataset Distribution}

While previous experiments used CIFAR-100 as the server-side public unlabeled dataset, a practical DART system should remain effective across different server data distributions. Table~\ref{table:server_dataset_results} shows that DART maintains consistent utility and robustness across diverse server datasets and sizes, with only marginal gains from larger or more varied datasets.

%% file: sec/7_conclusion.tex
\section{Conclusion}
\label{sec:conclusion}
We introduced DART, a server-side plug-in that enhances model robustness while preserving data privacy and incurring zero client-side overhead. Extensive experiments demonstrate that DART consistently improves the robustness of diverse FL algorithms. Its plug-in design enables seamless integration into existing FL systems for robust model training. Future works could extend DART to other OOD scenarios and to transformer-based models.


%% file: sec/X_suppl.tex
\clearpage
\setcounter{page}{1}
\maketitlesupplementary

\section{Code}
To allow for reproducibility of our results and any future works, our code will be made available upon acceptance.

\section{Method Details}

\subsection{Data Augmentation}
\label{appendix:data_augmentation}
The DART method adopts the standard AugMix~\cite{hendrycks2019augmix} augmentation strategy, which stochastically mixes diverse transformation operations such as posterize, equalize, rotate, translate, and shear. This procedure introduces multiple sources of randomness, including the selection of operations, their severity levels, the depth of transformation chains, and mixing weights. 

First, three operations, $\text{op}_1, \text{op}_2, \text{and } \text{op}_3$,  are sampled from the set of available operations, $\mathcal{O}$. The sampled operations are sequentially composed to generate transformation chains of increasing depth:

\begin{equation}
    \text{op}_{12} = \text{op}_2 \circ \text{op}_1\text{, and }\text{op}_{123} = \text{op}_3 \circ \text{op}_2 \circ \text{op}_1
\end{equation}

Then one chain is sampled uniformly at random from $\{\text{op}_1, \text{op}_{12}, \text{op}_{123}\}$. This process is repeated $S$ times, where $S$ is a hyperparameter denoting mixture width, resulting in $S$ operation chains $\{\text{chain}_i\}^{S}_{i=1}$. $S$ is set to 3 by default~\cite{hendrycks2019augmix}.

The augmented sample $\augx$ is generated from a sample $\x$ using: 
\begin{equation}
    \augx = \eta\x+(1-\eta)\sum_{i=1}^{S} m_i \times \text{chain}_i(\x)
\end{equation}
where $(m_1 \dots m_i)$ and $\eta$ are random weights sampled from $\text{Dirichlet}(\alpha \dots \alpha)$ and from $\text{Beta}(\alpha, \alpha)$ distributions, respectively. As can be seen, the augmentation procedure mixes the output of the operation chains. It also includes direct contribution from the original image $\x$ to prevent the loss of sematic information. 

This diverse augmentation process is designed to improve the robustness of machine learning models by exposing them to a wide range of data during training.

\subsection{Overall System Algorithm}
\label{appendix:system}
Section~\ref{sec:system} presented the overall DART-enhanced FL system design. In a standard FL setting, clients perform local training while a central server aggregates their updates. Our proposed plug-in, \textbf{DART}, is integrated on the server side, operating immediately after model aggregation and before the updated global model is redistributed to the clients. Algorithm~\ref{alg:system} details the DART-enhanced FL system.

\begin{algorithm}[h]
\caption{\texttt{System Description}. \\ $K$: number of clients; $D_k$: local private dataset of client $k$; $D_0$: public unlabeled dataset; $G$: number of global updates; $E$: local rounds per global update; $\robT$: global rounds per DART update}
\label{alg:system}
\textbf{Server executes:}
\begin{algorithmic}[1]
    \State Initialize $\w_0$;
    \For{each global round $t = 1, 2, \dots, G$} 
        \For{each client $k = 1, 2, \dots, K$ \textbf{in parallel}}
            \State $\w_{k}^{t+1} \gets \texttt{ClientUpdate}(k, \w_0^t)$;
        \EndFor
        \State $\w^{t+1}_0 \gets \texttt{Aggregate}(\w_{1}^{t+1}, \dots, \w_{k}^{t+1})$;
        \If{$t \bmod \robT = 0$}
            \State $\w^{t+1}_{\text{rob}} \gets \texttt{DART}(\w^{t+1}_0, D_0)$;
            \State $\w^{t+1}_0 \gets \w^{t+1}_{\text{rob}}$;
        \EndIf
    \EndFor
\end{algorithmic}
\textbf{ClientUpdate($k$, $\w$):}
\begin{algorithmic}[1]
    \For{each local epoch $i = 1, 2, \dots, E$}
        \For{batch $\batch \in D_k$}
            \State $\w \gets \texttt{LocalTraining}(\w, \batch)$;
        \EndFor
    \EndFor
    \State \textbf{return} $\w$ to server
\end{algorithmic}
\end{algorithm}

\subsection{DART Algorithm}
\label{appendix:dart}
Section~\ref{sec:DART} introduced our novel server-side plug-in, \textbf{DART}, designed to enable resource-efficient and robust federated learning. Algorithm~\ref{alg:DART} details the DART procedure.

\begin{algorithm}[h]
\caption{\texttt{DART}.\\ $\DARTD$ and $\valD$: public unlabeled training and validation datasets partitioned from $D_0$; $\maxT$: maximum number of DART epochs; $\valT$: early stopping threshold; $\DARTeta$: DART learning rate; $\w_0$: pre-trained weights.}
\label{alg:DART}
\begin{algorithmic}[1]
\State counter $\gets 0$; $\w \gets \w_0$; $\text{loss}_\text{min} \gets \infty$
\For{each epoch $i = 1, 2, \dots, \maxT$}
    \For{batch $\batch \in \DARTD$}
        \State $\w \gets \w - \DARTeta \nabla \DARTloss(\w; \batch)$
    \EndFor
    \If{$\DARTloss(\w; \valD) \leq \text{loss}_\text{min}$}
        \State $\text{loss}_\text{min} \gets \DARTloss(\w; \valD)$
        \State $\w_\text{best} \gets \w$; counter $\gets 0$
    \Else
        \State counter \texttt{++}
    \EndIf
    \If{counter $= \valT - 1$}
        \State break
    \EndIf
\EndFor
\State $\DARTw \gets \w_\text{best}$
\State \textbf{return} $\DARTw$
\end{algorithmic}
\end{algorithm}

\section{Theorem Proof}
\label{appendix:proof}
The theoretical analysis in this work is inspired by~\citet{lin2020ensemble}, who study ensemble distillation for model fusion in model heterogeneous FL. In their work, a generalization bound is developed that relates the loss of distilled federated models to that of a centrally trained model, explicitly characterizing the impact of distribution mismatch across clients and server. Building on this foundation, we derive a bound that relates the pre-DART (teacher) and post-DART (student) clean risks, enabling us to formally quantify the utility degradation introduced by DART while enhancing robustness.

\paragraph{Setup and notation.}
Let $\mathcal{X}$ be the input space and $\mathcal{Y}$ the label set.
Let $\inD$ and $\outD$ be distributions on $\mathcal{X}\times\mathcal{Y}$.
A classifier $h:\mathcal{X}\to\mathcal{Y}$ has risk on a distribution $Q$
\begin{equation}
R_Q(h) \;=\; \Pr_{(\x,\y)\sim Q}\big(h(\x)\neq \y\big)
\end{equation}
For $h,h':\mathcal{X}\to\mathcal{Y}$, define their \emph{disagreement probability} on a distribution $Q$ by
\begin{equation}
\mathrm{Dis}_{Q}(h,h') \;=\; \Pr_{\x\sim Q}\big(h(\x)\neq h'(\x)\big)
\end{equation}

We consider a teacher $f_{\w_\text{t}}$ and a student $f_{\w_\text{s}}$ that make predictions via posteriors
$p_\text{t}(\cdot\mid \x) \text{ and }p_\text{s}(\cdot\mid \x)$, with decisions
$f_{\w_\text{t}}(\x)=\arg\max_\y p_\text{t}(\y\mid \x)$ and $f_{\w_\text{s}}(\x)=\arg\max_\y p_\text{s}(\y\mid \x)$. The client distribution is $\inD$, while $\outD$ is the server distribution.

Define the \emph{teacher margin} at $\x$:
\begin{equation}
m_\text{t}(\x)\;=\;p_\text{t}(\y_\text{t}^*\mid \x)-\max_{\y\neq \y_\text{t}^*}p_\text{t}(\y\mid \x)
\quad
\end{equation}
where $\quad \y_\text{t}^*=\arg\max_\y p_t(\y\mid x)$ and $m_\text{t}(\x)\in[0,1]$. Define the \emph{margin tail} on $\outD$ by
$\theta_{\outD}(\gamma)=\Pr_{\x\sim \outD}\big(m_\text{t}(\x)\le \gamma\big)$ for $\gamma\in(0,1]$.
The total variation distance is 
\begin{equation}
\mathrm{TV}(\x)\;=\;\tfrac12\big\|p_\text{t}(\cdot\mid \x)-p_\text{s}(\cdot\mid \x)\big\|_1
\end{equation}
and the expected distillation KL divergence on $\outD$:
\begin{equation}
\delta \;=\; \mathbb{E}_{\x\sim \outD}\Big[\mathrm{KL}\!\big(p_\text{t}(\cdot\mid \x)\,\|\,p_\text{s}(\cdot\mid \x)\big)\Big]
\end{equation}
Finally, the $\mathcal{H}\Delta\mathcal{H}$-divergence~\cite{ben2010theory} between $\inD$ and $\outD$ is
\[
d_{\mathcal{H}\Delta\mathcal{H}}(\inD,\outD)
\;=\; 2\sup_{h,h'\in\mathcal{H}}
\Big|\mathrm{Dis}_{\inD}(h,h')-\mathrm{Dis}_{\outD}(h,h')\Big|
\]
for some hypothesis class $\mathcal{H}$.

\begin{lemma}[]
\label{lem:flip-tv}
For any $\x\in\mathcal{X}$, if $f_{\w_\text{s}}\neq f_{\w_\text{t}}$ then
\[
m_\text{t}(\x)\;\le\;2\,\mathrm{TV}(\x)
\]
\end{lemma}

\begin{proof}
Let $\y_\text{t}^*=f_{\w_\text{t}}(\x)$ and let $\y'=f_s(x)\neq \y_t^*$. Since the student prefers $\y'$,
$p_\text{s}(\y'\mid \x)\ge p_\text{s}(\y_t^*\mid \x)$. Let $\Delta_\y:=p_\text{s}(\y\mid \x)-p_\text{t}(\y\mid \x)$ for each class $\y$, then


\begin{align}
m_\text{t}(\x)
&\leq p_\text{t}(\y_t^*\mid \x)-p_\text{t}(\y'\mid \x) \\
&= \begin{aligned}[t]
   \big(p_\text{t}(&\y_\text{t}^*\mid \x)-p_\text{s}(\y_\text{t}^*\mid \x)\big)
 \\
 &+ \big(p_\text{s}(\y_\text{t}^*\mid x)-p_\text{s}(\y'\mid \x)\big)\\
 &+ \big(p_\text{s}(\y'\mid \x)-p_\text{t}(\y'\mid \x)\big)
\end{aligned} \\
&\le |\Delta_{\y_\text{t}^*}| + 0 + |\Delta_{\y'}| \\
&\le \sum_{\y\in\mathcal{Y}} |\Delta_\y| \\
&= \big\|p_\text{s}(\cdot\mid \x)-p_\text{t}(\cdot\mid \x)\big\|_1 \\
&= 2\,\mathrm{TV}(\x)
\end{align}
which proves the claim.

\end{proof}

\begin{theorem*}[Clean Student Risk Bound under Distillation]
\label{thm:main}
For any $\gamma\in(0,1]$,
\begin{align}
    \R_{\inD}(f_{\w_\text{s}})
    &\le
    \R_{\inD}(f_{\w_\text{t}})
    + \theta_{\outD}(\gamma) \nonumber \\
    &\quad
    + \frac{2}{\gamma}\sqrt{\frac{\delta}{2}}
    + \frac{1}{2}\, d_{\mathcal{H}\Delta\mathcal{H}}(\inD, \outD)
    \label{eq:main_appdx}
\end{align}

\end{theorem*}

\begin{proof}
\textbf{(1) Risk decomposition.}
For any $(\x,\y)$,
\begin{equation}
\mathbf{1}\big[f_{\w_\text{s}}(\x)\neq \y\big]
\;\le\;
\mathbf{1}\big[f_{\w_\text{t}}(\x)\neq \y\big] + \mathbf{1}\big[f_{\w_\text{s}}(\x)\neq f_{\w_\text{t}}(\x)\big]
\end{equation}
Taking expectation over $(\x,\y)\sim \inD$ gives
\begin{equation}
\label{eq:decomp}
R_{\inD}(f_{\w_\text{s}})
\;\le\;
R_{\inD}(f_{\w_\text{t}}) + \mathrm{Dis}_{\inD}(f_{\w_\text{s}},f_{\w_\text{t}})
\end{equation}

\medskip
\noindent\textbf{(2) Move disagreement from $\inD$ to $\outD$.}
By the definition of $d_{\mathcal{H}\Delta\mathcal{H}}$,
\begin{equation}
\mathrm{Dis}_{\inD}(f_{\w_\text{s}},f_{\w_\text{t}})
\;\le\;
\mathrm{Dis}_{\outD}(f_{\w_\text{s}},f_{\w_\text{t}})
\;+\;
\frac{1}{2}\,d_{\mathcal{H}\Delta\mathcal{H}}(\inD,\outD)
\end{equation}
Combining with \eqref{eq:decomp},
\begin{equation}
\label{eq:after-divergence}
\begin{aligned}
R_{\inD}(f_{\w_\text{s}})
&\le
R_{\inD}(f_{\w_\text{t}})
+ \mathrm{Dis}_{\outD}(f_{\w_\text{s}},f_{\w_\text{t}})\\
&\quad
+ \frac{1}{2}\,d_{\mathcal{H}\Delta\mathcal{H}}(\inD,\outD)
\end{aligned}
\end{equation}

\medskip
\noindent\textbf{(3) Control disagreement on $\outD$ via margin and TV.}
By Lemma~\ref{lem:flip-tv}, for any $\x$,
\[
\big\{f_{\w_\text{s}}(\x)\neq f_{\w_\text{t}}(\x)\big\}\;\subseteq\;\big\{m_\text{t}(\x)\le 2\,\mathrm{TV}(\x)\big\}
\]
Fix $\gamma\in(0,1]$. Then
$\{m_\text{t}\le 2\,\mathrm{TV}\}\subseteq \{m_\text{t}\le \gamma\}\cup\{2\,\mathrm{TV}\ge \gamma\}$ and,
\[
\mathrm{Dis}_{\outD}(f_{\w_\text{s}},f_{\w_\text{t}})
\;\le\;
\theta_{\outD}(\gamma)
\;+\;
\Pr_{\x\sim \outD}\big(\,2\,\mathrm{TV}(\x)\ge \gamma\,\big)
\]
Now apply Markov's inequality (since $\mathrm{TV}\ge 0$), Pinsker and Jensen:
\begin{align}
\Pr(\,2\,\mathrm{TV}\ge \gamma\,)
\;&\le\;
\frac{2}{\gamma}\,\mathbb{E}[\mathrm{TV}]
\;\\&\le\;
\frac{2}{\gamma}\,\mathbb{E}\Big[\sqrt{\tfrac12\,\mathrm{KL}(p_\text{t}\|p_\text{s})}\Big]
\;\\&\le\;
\frac{2}{\gamma}\sqrt{\tfrac12\,\mathbb{E}[\mathrm{KL}(p_\text{t}\|p_\text{s})]}
\;=\;
\frac{2}{\gamma}\sqrt{\frac{\delta}{2}}
\end{align}
Hence
\begin{equation}
\label{eq:d0-dis}
\mathrm{Dis}_{\outD}(f_{\w_\text{s}},f_{\w_\text{t}})
\;\le\;
\theta_{\outD}(\gamma)
\;+\;
\frac{2}{\gamma}\sqrt{\frac{\delta}{2}}
\end{equation}

\medskip
\noindent\textbf{(4) Combine.}
Insert \eqref{eq:d0-dis} into \eqref{eq:after-divergence} to obtain the stated bound.
\end{proof}

\section{Resource Setup}
\label{appendix:resource}
\subsection{Profiling Client Hardware}
 Each client is represented by an NVIDIA Jetson Orin Nano~\cite{nvidiaJetsonNano}, equipped with an Ampere GPU, a 6-core ARM CPU, and 8 GB of RAM, delivering up to 67 TOPS of compute. We profile batch execution time and batch energy consumption directly on the hardware. Power is measured using Jetson Stats~\cite{jetson-stats}, and energy is computed by integrating power over time, approximated as the product of sampled power values and their corresponding time intervals. Execution time is measured as wall-clock time. Table~\ref{tab:resources} reports the measured batch time and energy. While our experiments are executed on NVIDIA L40S GPUs, we use measurements collected on the Jetson Orin Nano to estimate the per-client training time and energy consumption for large-scale federated learning.

\begin{table}[h]
\centering
\caption{Client-side batch time and batch energy, measured on NVIDIA Jetson Orin Nano using ResNet-18 with batch size 64 and input dimension $32\times32\times3$.}
\resizebox{\linewidth}{!}{%
\begin{tabular}{lccccc}
\toprule
 & \multicolumn{2}{c}{Clean FL} & \multicolumn{3}{c}{Robust FL} \\
 \cmidrule(lr){2-3}
 \cmidrule(lr){4-6}
 & FedAvg & FedProx  & FedAFA & FedAugMix & FedPrime \\
\midrule
Batch Time (s)        
& 0.39 & 0.41 & 0.75 & 1.02 & 8.28 \\

Batch Energy (J)   
& 1.43 & 1.53 & 2.76 & 4.15 & 8.20 \\
\bottomrule
\end{tabular}%
}
\label{tab:resources}
\end{table}

 \subsection{Time Model}
 Since multiple clients perform training in parallel during each global round, the training time at round $i$ can be estimated as

 \begin{equation}
     \trainTime^i = i \times t_\text{batch}  \times b_\text{avg}  
 \end{equation}
where $t_\text{batch}$ is the profiled batch execution time and $ b_\text{avg}$ is the average number of batches processed by a client. $b_\text{avg}$ is computed as the average number of samples per client divided by the batch size:
 
 \begin{equation}
 b_\text{avg} = \frac{\sum_{k=1}^K |D_k|}{K\times\text{batch\_size}}
 \label{eq:bavg}
 \end{equation}
 where $K$ is the number of clients and $|D_k|$ is denotes the number of samples held by client $k$.

 \subsection{Energy Model}
 We consider the total compute energy consumed by clients during federated training, which can be estimated for global round $i$ as

 \begin{equation}
     \trainEnergy^i = i \times r\times K \times e_\text{batch}  \times b_\text{avg}  
 \end{equation}
where $e_\text{batch}$ denotes the profiled per-batch energy consumption, $r$ denotes the client participation rate, $K$ is the number of clients, and $ b_\text{avg}$ is computed using Eq.~\ref{eq:bavg}.

\section{CIFAR-10-C Results}
\subsection{CIFAR-10-C Corruptions}
CIFAR-10-C~\cite{hendrycks2019benchmarking} is a standard benchmark for evaluating model robustness to common image corruptions. Table~\ref{tab:cifar10c_corruptions} lists the corruption types included in the benchmark along with the abbreviations used throughout the paper.

\begin{table}[h]
    \centering
    \caption{CIFAR-10-C corruption types and abbreviations used throughout the paper.}
    \begin{tabular}{ll}
        \hline
        Corruption type & Abbreviation \\
        \hline
        Brightness         & bright \\
        Contrast           & contr  \\
        Defocus blur       & defoc  \\
        Elastic transform  & elas   \\
        Fog                & fog    \\
        Frost              & frost  \\
        Gaussian blur      & g\_blur  \\
        Gaussian noise     & g\_noise \\
        Glass blur         & glass  \\
        Impulse noise      & imp    \\
        JPEG compression   & jpeg   \\
        Motion blur        & motion \\
        Pixelate           & pix    \\
        Saturate           & sat    \\
        Shot noise         & shot   \\
        Snow               & snow   \\
        Spatter            & spatt  \\
        Speckle noise      & speck  \\
        Zoom blur          & zoom   \\
        \hline
    \end{tabular}
    
    \label{tab:cifar10c_corruptions}
\end{table}

\subsection{Full CIFAR-10-C Results}
 Section~\ref{sec:DARTvsClean} evaluates popular FL methods and their DART-enhanced variants on the CIFAR-10-C robustness benchmark. Table~\ref{tab:full_plugin_result} extends the results by reporting performance across all corruption types. As discussed earlier, integrating DART consistently improves robustness on CIFAR-10-C at zero client-side overhead and at varying degrees of heterogeneity. Notably, DART improves performance across most corruption categories, with the exception of brightness and saturation, which introduce minimal visual distortion and thus exhibit smaller robustness gains.

\begin{table*}[t]
\centering
\caption{Impact of DART in enhancing the robustness of diverse standard FL methods. Clean accuracy is measured on CIFAR-10 ($\clnA$), and robust accuracy on CIFAR-10-C ($\robA^\mathrm{C}$) for ResNet-18. Accuracies on all CIFAR-10-C corruptions are shown. 
DART significantly enhances the robustness (avg. $4.3\%$) of standard FL methods while incurring a small drop (avg. $1.6\%$) in clean accuracy at no  additional computational cost for the clients.}
\resizebox{\textwidth}{!}{%
{\Huge
\begin{tabular}{lccccccccccccccccccccc}
\toprule
Method & $\clnA (\%) $ & $\robA^\mathrm{C} (\%)$ & bright $(\%)$ & contr $(\%)$ & defoc $(\%)$ & elas $(\%)$ & fog $(\%)$ & frost $(\%)$ & g\_blur $(\%)$ & g\_noise $(\%)$ & glass $(\%)$ & imp $(\%)$ & jpeg $(\%)$ & motion $(\%)$ & pix $(\%)$ & sat $(\%)$ & shot $(\%)$ & snow $(\%)$ & spatt $(\%)$ & speck $(\%)$ & zoom $(\%)$\\
\midrule
\multicolumn{22}{c}{$\alpha_\text{IID} = 100$} \\
\midrule

FedAvg & 90.3 \Large $\pm$ 0.9 & 70.5 \Large $\pm$ 0.5 & 88.9 \Large $\pm$ 0.9 & 68.2 \Large $\pm$ 2.3 & 75.8 \Large $\pm$ 2.4 & 79.2 \Large $\pm$ 1.3 & 82.5 \Large $\pm$ 1.4 & 72.2 \Large $\pm$ 1.7 & 67.6 \Large $\pm$ 3.2 & 47.8 \Large $\pm$ 6.6 & 46.0 \Large $\pm$ 3.3 & 55.6 \Large $\pm$ 4.2 & 77.7 \Large $\pm$ 2.1 & 71.8 \Large $\pm$ 3.7 & 73.2 \Large $\pm$ 2.5 & 87.1 \Large $\pm$ 1.1 & 58.3 \Large $\pm$ 4.8 & 75.2 \Large $\pm$ 1.3 & 79.6 \Large $\pm$ 1.1 & 60.6 \Large $\pm$ 4.1 & 71.6 \Large $\pm$ 2.4  \\

FedDyn & 85.0 \Large $\pm$ 1.0 & 67.7 \Large $\pm$ 1.3 & 83.1 \Large $\pm$ 1.1 & 63.7 \Large $\pm$ 2.4 & 73.7 \Large $\pm$ 2.9 & 74.6 \Large $\pm$ 2.1 & 76.9 \Large $\pm$ 2.0 & 68.1 \Large $\pm$ 2.7 & 66.9 \Large $\pm$ 3.1 & 53.5 \Large $\pm$ 6.2 & 43.3 \Large $\pm$ 6.9 & 56.9 \Large $\pm$ 2.6 & 75.8 \Large $\pm$ 1.1 & 65.9 \Large $\pm$ 4.2 & 66.0 \Large $\pm$ 2.0 & 81.8 \Large $\pm$ 1.2 & 61.3 \Large $\pm$ 5.0 & 68.4 \Large $\pm$ 2.0 & 75.5 \Large $\pm$ 1.1 & 62.5 \Large $\pm$ 4.6 & 68.3 \Large $\pm$ 3.4  \\

FedNova & 86.5 \Large $\pm$ 0.5 & 69.5 \Large $\pm$ 0.8 & 84.5 \Large $\pm$ 0.4 & 61.9 \Large $\pm$ 0.9 & 75.0 \Large $\pm$ 0.4 & 76.6 \Large $\pm$ 0.4 & 76.4 \Large $\pm$ 0.4 & 69.1 \Large $\pm$ 1.3 & 68.4 \Large $\pm$ 0.9 & 56.4 \Large $\pm$ 2.0 & 45.6 \Large $\pm$ 2.4 & 58.9 \Large $\pm$ 0.9 & 77.9 \Large $\pm$ 0.7 & 68.4 \Large $\pm$ 0.9 & 71.3 \Large $\pm$ 2.0 & 83.5 \Large $\pm$ 0.7 & 63.1 \Large $\pm$ 1.6 & 70.9 \Large $\pm$ 0.5 & 78.6 \Large $\pm$ 0.6 & 63.7 \Large $\pm$ 1.5 & 70.2 \Large $\pm$ 0.9  \\

FedProx & \textbf{91.4} \Large $\pm$ 0.3 & 72.7 \Large $\pm$ 1.0 & \textbf{89.9} \Large $\pm$ 0.4 & 73.5 \Large $\pm$ 1.6 & 79.1 \Large $\pm$ 1.0 & 81.0 \Large $\pm$ 1.0 & \textbf{84.7} \Large $\pm$ 0.7 & 73.7 \Large $\pm$ 1.3 & 70.1 \Large $\pm$ 1.8 & 51.7 \Large $\pm$ 2.8 & 45.5 \Large $\pm$ 2.5 & 60.7 \Large $\pm$ 2.1 & 79.4 \Large $\pm$ 0.7 & 73.8 \Large $\pm$ 1.1 & 71.5 \Large $\pm$ 2.3 & \textbf{88.5} \Large $\pm$ 0.4 & 61.8 \Large $\pm$ 2.5 & 77.2 \Large $\pm$ 1.1 & 82.4 \Large $\pm$ 1.0 & 64.0 \Large $\pm$ 2.3 & 73.5 \Large $\pm$ 1.5  \\

\cmidrule(lr){1-22}

FedAvg+DART & 89.2 \Large $\pm$ 0.6 & 77.4 \Large $\pm$ 0.9 & 87.4 \Large $\pm$ 0.9 & 73.8 \Large $\pm$ 2.1 & 84.8 \Large $\pm$ 1.2 & 81.2 \Large $\pm$ 1.3 & 82.5 \Large $\pm$ 1.9 & 75.2 \Large $\pm$ 0.9 & 82.6 \Large $\pm$ 1.5 & 64.0 \Large $\pm$ 2.3 & 57.7 \Large $\pm$ 2.1 & 69.2 \Large $\pm$ 1.8 & 79.2 \Large $\pm$ 1.1 & 80.1 \Large $\pm$ 2.3 & \textbf{77.8} \Large $\pm$ 1.8 & 86.2 \Large $\pm$ 0.6 & 71.7 \Large $\pm$ 1.5 & 77.2 \Large $\pm$ 1.2 & 82.8 \Large $\pm$ 1.1 & 73.8 \Large $\pm$ 1.5 & 82.5 \Large $\pm$ 1.4  \\

FedDyn+DART & 83.2 \Large $\pm$ 1.8 & 71.2 \Large $\pm$ 1.2 & 81.2 \Large $\pm$ 1.3 & 67.4 \Large $\pm$ 1.9 & 78.6 \Large $\pm$ 1.6 & 74.8 \Large $\pm$ 1.2 & 75.9 \Large $\pm$ 1.2 & 67.9 \Large $\pm$ 1.4 & 76.2 \Large $\pm$ 1.2 & 59.2 \Large $\pm$ 3.0 & 51.5 \Large $\pm$ 4.7 & 65.2 \Large $\pm$ 1.4 & 74.6 \Large $\pm$ 3.7 & 72.8 \Large $\pm$ 2.2 & 70.3 \Large $\pm$ 7.6 & 80.2 \Large $\pm$ 2.0 & 66.2 \Large $\pm$ 2.0 & 69.9 \Large $\pm$ 1.6 & 76.8 \Large $\pm$ 2.6 & 68.0 \Large $\pm$ 1.9 & 75.7 \Large $\pm$ 0.7  \\

FedNova+DART & 85.1 \Large $\pm$ 0.6 & 72.6 \Large $\pm$ 0.8 & 82.8 \Large $\pm$ 0.5 & 66.7 \Large $\pm$ 1.4 & 81.0 \Large $\pm$ 0.8 & 77.5 \Large $\pm$ 0.8 & 76.1 \Large $\pm$ 0.9 & 68.1 \Large $\pm$ 1.1 & 78.7 \Large $\pm$ 0.9 & 60.0 \Large $\pm$ 1.0 & 51.0 \Large $\pm$ 1.6 & 64.5 \Large $\pm$ 1.0 & 77.2 \Large $\pm$ 0.8 & 74.3 \Large $\pm$ 1.4 & 76.2 \Large $\pm$ 0.7 & 82.3 \Large $\pm$ 0.6 & 67.0 \Large $\pm$ 0.8 & 70.9 \Large $\pm$ 0.8 & 78.8 \Large $\pm$ 0.6 & 68.6 \Large $\pm$ 0.9 & 78.1 \Large $\pm$ 0.9  \\

FedProx+DART & 90.2 \Large $\pm$ 0.4 & \textbf{79.0} \Large $\pm$ 0.8 & 88.6 \Large $\pm$ 0.6 & \textbf{77.7} \Large $\pm$ 2.0 & \textbf{86.1} \Large $\pm$ 0.7 & \textbf{82.4} \Large $\pm$ 0.6 & 84.2 \Large $\pm$ 1.4 & \textbf{77.3} \Large $\pm$ 1.4 & \textbf{83.9} \Large $\pm$ 0.7 & \textbf{65.6} \Large $\pm$ 2.7 & \textbf{57.9} \Large $\pm$ 3.6 & \textbf{72.6} \Large $\pm$ 1.9 & \textbf{80.9} \Large $\pm$ 0.4 & \textbf{81.8} \Large $\pm$ 1.0 & 77.7 \Large $\pm$ 1.3 & 87.3 \Large $\pm$ 0.8 & \textbf{73.7} \Large $\pm$ 1.9 & \textbf{79.2} \Large $\pm$ 1.0 & \textbf{84.8} \Large $\pm$ 0.8 & \textbf{76.0} \Large $\pm$ 1.5 & \textbf{83.5} \Large $\pm$ 0.8  \\

\midrule
\multicolumn{22}{c}{$\alpha_\text{IID} = 10$} \\
\midrule

FedAvg & 89.5 \Large $\pm$ 0.9 & 69.6 \Large $\pm$ 1.8 & 88.1 \Large $\pm$ 0.9 & 68.6 \Large $\pm$ 1.7 & 76.0 \Large $\pm$ 1.0 & 78.3 \Large $\pm$ 1.5 & 82.4 \Large $\pm$ 0.8 & 69.7 \Large $\pm$ 2.5 & 68.0 \Large $\pm$ 1.1 & 46.6 \Large $\pm$ 7.4 & 43.6 \Large $\pm$ 3.2 & 54.3 \Large $\pm$ 5.9 & 77.2 \Large $\pm$ 2.0 & 72.6 \Large $\pm$ 1.1 & 72.2 \Large $\pm$ 4.1 & 86.4 \Large $\pm$ 1.3 & 56.6 \Large $\pm$ 6.3 & 74.1 \Large $\pm$ 1.9 & 78.8 \Large $\pm$ 1.5 & 58.4 \Large $\pm$ 6.0 & 71.1 \Large $\pm$ 2.2  \\

FedDyn & 83.9 \Large $\pm$ 1.7 & 66.7 \Large $\pm$ 1.1 & 82.0 \Large $\pm$ 1.8 & 60.9 \Large $\pm$ 4.0 & 72.6 \Large $\pm$ 1.8 & 72.6 \Large $\pm$ 2.0 & 75.1 \Large $\pm$ 2.3 & 65.9 \Large $\pm$ 1.9 & 66.5 \Large $\pm$ 1.7 & 52.5 \Large $\pm$ 6.6 & 43.5 \Large $\pm$ 6.8 & 57.2 \Large $\pm$ 2.4 & 73.9 \Large $\pm$ 1.8 & 66.6 \Large $\pm$ 0.6 & 67.1 \Large $\pm$ 5.1 & 80.9 \Large $\pm$ 1.5 & 59.7 \Large $\pm$ 4.9 & 67.4 \Large $\pm$ 1.5 & 74.9 \Large $\pm$ 1.7 & 60.6 \Large $\pm$ 4.2 & 67.1 \Large $\pm$ 1.7  \\

FedNova & 85.7 \Large $\pm$ 0.3 & 67.9 \Large $\pm$ 0.5 & 83.6 \Large $\pm$ 0.4 & 60.1 \Large $\pm$ 1.3 & 73.6 \Large $\pm$ 0.9 & 75.1 \Large $\pm$ 0.7 & 74.7 \Large $\pm$ 1.1 & 67.1 \Large $\pm$ 0.5 & 66.5 \Large $\pm$ 1.2 & 54.2 \Large $\pm$ 1.6 & 44.6 \Large $\pm$ 1.0 & 57.3 \Large $\pm$ 1.8 & 76.8 \Large $\pm$ 0.6 & 65.6 \Large $\pm$ 0.7 & 70.0 \Large $\pm$ 1.7 & 82.7 \Large $\pm$ 0.6 & 61.1 \Large $\pm$ 1.6 & 69.2 \Large $\pm$ 0.3 & 77.7 \Large $\pm$ 0.7 & 61.7 \Large $\pm$ 1.5 & 68.4 \Large $\pm$ 1.1  \\

FedProx & \textbf{90.6} \Large $\pm$ 0.4 & 71.9 \Large $\pm$ 1.0 & \textbf{89.2} \Large $\pm$ 0.5 & 71.9 \Large $\pm$ 2.1 & 79.4 \Large $\pm$ 1.0 & 80.6 \Large $\pm$ 0.5 & \textbf{83.6} \Large $\pm$ 1.1 & 71.7 \Large $\pm$ 0.9 & 71.2 \Large $\pm$ 2.1 & 50.0 \Large $\pm$ 4.2 & 45.6 \Large $\pm$ 3.1 & 59.1 \Large $\pm$ 2.4 & 77.8 \Large $\pm$ 0.8 & 73.9 \Large $\pm$ 1.1 & 71.2 \Large $\pm$ 1.0 & \textbf{87.9} \Large $\pm$ 0.2 & 59.7 \Large $\pm$ 3.1 & 75.6 \Large $\pm$ 0.9 & 82.0 \Large $\pm$ 1.0 & 61.6 \Large $\pm$ 2.6 & 74.9 \Large $\pm$ 1.5  \\

\cmidrule(lr){1-22}

FedAvg+DART & 88.3 \Large $\pm$ 0.7 & 75.8 \Large $\pm$ 1.5 & 86.4 \Large $\pm$ 1.2 & 71.8 \Large $\pm$ 1.0 & 83.4 \Large $\pm$ 1.2 & 79.5 \Large $\pm$ 1.3 & 81.6 \Large $\pm$ 0.9 & 73.2 \Large $\pm$ 2.6 & 81.1 \Large $\pm$ 1.5 & 62.1 \Large $\pm$ 3.9 & 54.7 \Large $\pm$ 4.1 & 67.4 \Large $\pm$ 2.4 & 78.0 \Large $\pm$ 0.8 & 78.9 \Large $\pm$ 1.7 & \textbf{77.6} \Large $\pm$ 1.0 & 85.7 \Large $\pm$ 0.8 & 70.0 \Large $\pm$ 3.1 & 75.2 \Large $\pm$ 1.7 & 81.2 \Large $\pm$ 1.4 & 72.1 \Large $\pm$ 2.8 & 81.1 \Large $\pm$ 1.3  \\

FedDyn+DART & 82.7 \Large $\pm$ 1.4 & 69.5 \Large $\pm$ 2.4 & 80.0 \Large $\pm$ 2.1 & 64.1 \Large $\pm$ 3.1 & 77.9 \Large $\pm$ 1.8 & 73.8 \Large $\pm$ 1.8 & 74.1 \Large $\pm$ 2.6 & 64.7 \Large $\pm$ 3.8 & 75.6 \Large $\pm$ 2.0 & 56.8 \Large $\pm$ 4.4 & 47.8 \Large $\pm$ 4.1 & 63.0 \Large $\pm$ 3.5 & 73.6 \Large $\pm$ 2.3 & 71.4 \Large $\pm$ 3.2 & 70.8 \Large $\pm$ 1.9 & 80.1 \Large $\pm$ 1.4 & 63.7 \Large $\pm$ 3.9 & 67.6 \Large $\pm$ 2.9 & 76.0 \Large $\pm$ 1.9 & 65.5 \Large $\pm$ 3.7 & 74.6 \Large $\pm$ 1.8  \\

FedNova+DART & 84.7 \Large $\pm$ 0.3 & 71.8 \Large $\pm$ 0.4 & 82.3 \Large $\pm$ 0.2 & 66.7 \Large $\pm$ 0.5 & 80.4 \Large $\pm$ 0.5 & 76.5 \Large $\pm$ 0.5 & 76.0 \Large $\pm$ 0.4 & 67.3 \Large $\pm$ 0.7 & 78.1 \Large $\pm$ 0.6 & 58.2 \Large $\pm$ 1.6 & 48.9 \Large $\pm$ 1.4 & 64.0 \Large $\pm$ 0.4 & 76.3 \Large $\pm$ 0.6 & 74.2 \Large $\pm$ 0.6 & 75.1 \Large $\pm$ 1.3 & 82.0 \Large $\pm$ 0.4 & 65.5 \Large $\pm$ 1.2 & 69.8 \Large $\pm$ 0.5 & 78.1 \Large $\pm$ 0.4 & 67.3 \Large $\pm$ 1.2 & 77.2 \Large $\pm$ 0.6  \\

FedProx+DART & 88.7 \Large $\pm$ 0.6 & \textbf{77.2} \Large $\pm$ 2.0 & 87.0 \Large $\pm$ 1.3 & \textbf{74.7} \Large $\pm$ 3.7 & \textbf{84.6} \Large $\pm$ 0.8 & 80.5 \Large $\pm$ 1.0 & 82.1 \Large $\pm$ 2.5 & \textbf{74.5} \Large $\pm$ 3.4 & \textbf{82.4} \Large $\pm$ 1.0 & \textbf{64.5} \Large $\pm$ 4.3 & \textbf{56.8} \Large $\pm$ 4.3 & \textbf{70.7} \Large $\pm$ 2.5 & \textbf{78.0} \Large $\pm$ 2.6 & \textbf{80.0} \Large $\pm$ 2.0 & 76.6 \Large $\pm$ 2.4 & 86.3 \Large $\pm$ 0.7 & \textbf{72.0} \Large $\pm$ 3.7 & \textbf{76.7} \Large $\pm$ 2.7 & \textbf{83.1} \Large $\pm$ 1.0 & \textbf{74.1} \Large $\pm$ 3.6 & \textbf{82.1} \Large $\pm$ 1.2  \\

\midrule
\multicolumn{22}{c}{$\alpha_\text{IID} = 1$} \\
\midrule

FedAvg & 87.8 \Large $\pm$ 1.4 & 65.6 \Large $\pm$ 1.9 & 85.7 \Large $\pm$ 1.0 & 63.8 \Large $\pm$ 1.8 & 73.0 \Large $\pm$ 1.4 & 73.5 \Large $\pm$ 2.1 & 77.3 \Large $\pm$ 2.4 & 65.4 \Large $\pm$ 4.0 & 65.2 \Large $\pm$ 1.8 & 42.1 \Large $\pm$ 4.8 & 39.2 \Large $\pm$ 4.3 & 50.1 \Large $\pm$ 3.1 & 71.9 \Large $\pm$ 2.4 & 65.4 \Large $\pm$ 2.7 & 70.7 \Large $\pm$ 4.8 & 84.4 \Large $\pm$ 1.3 & 52.3 \Large $\pm$ 5.0 & 69.6 \Large $\pm$ 2.6 & 75.1 \Large $\pm$ 1.5 & 54.5 \Large $\pm$ 5.0 & 66.4 \Large $\pm$ 2.5  \\

FedDyn & 80.7 \Large $\pm$ 1.0 & 64.3 \Large $\pm$ 1.1 & 78.4 \Large $\pm$ 2.1 & 58.9 \Large $\pm$ 2.0 & 69.9 \Large $\pm$ 2.3 & 70.7 \Large $\pm$ 2.2 & 71.8 \Large $\pm$ 1.7 & 62.6 \Large $\pm$ 2.8 & 63.9 \Large $\pm$ 3.1 & 52.7 \Large $\pm$ 4.1 & 44.1 \Large $\pm$ 5.6 & 53.9 \Large $\pm$ 2.5 & 72.7 \Large $\pm$ 1.2 & 62.6 \Large $\pm$ 3.9 & 63.0 \Large $\pm$ 3.0 & 77.0 \Large $\pm$ 1.5 & 58.8 \Large $\pm$ 3.3 & 64.4 \Large $\pm$ 1.7 & 71.8 \Large $\pm$ 1.4 & 59.2 \Large $\pm$ 2.9 & 65.2 \Large $\pm$ 3.6  \\

FedNova & 82.7 \Large $\pm$ 0.6 & 64.9 \Large $\pm$ 0.8 & 80.3 \Large $\pm$ 0.5 & 55.3 \Large $\pm$ 0.8 & 69.3 \Large $\pm$ 0.9 & 70.4 \Large $\pm$ 0.8 & 69.0 \Large $\pm$ 1.2 & 64.3 \Large $\pm$ 0.4 & 62.8 \Large $\pm$ 1.0 & 54.1 \Large $\pm$ 1.6 & 45.6 \Large $\pm$ 1.7 & 55.3 \Large $\pm$ 1.3 & 74.2 \Large $\pm$ 0.9 & 60.1 \Large $\pm$ 1.3 & 68.3 \Large $\pm$ 1.0 & 79.5 \Large $\pm$ 0.5 & 60.2 \Large $\pm$ 1.4 & 66.4 \Large $\pm$ 0.9 & 74.9 \Large $\pm$ 1.0 & 60.4 \Large $\pm$ 1.4 & 63.4 \Large $\pm$ 0.7  \\

FedProx & \textbf{88.9} \Large $\pm$ 1.1 & 70.2 \Large $\pm$ 2.0 & \textbf{86.8} \Large $\pm$ 0.8 & 68.1 \Large $\pm$ 1.5 & 77.1 \Large $\pm$ 2.1 & \textbf{77.4} \Large $\pm$ 1.5 & \textbf{80.2} \Large $\pm$ 1.2 & 69.4 \Large $\pm$ 2.9 & 69.3 \Large $\pm$ 3.0 & 51.7 \Large $\pm$ 5.9 & 42.9 \Large $\pm$ 5.7 & 58.7 \Large $\pm$ 5.0 & 76.6 \Large $\pm$ 1.7 & 70.0 \Large $\pm$ 3.4 & 71.4 \Large $\pm$ 2.2 & \textbf{85.9} \Large $\pm$ 1.1 & 60.8 \Large $\pm$ 4.7 & \textbf{72.9} \Large $\pm$ 1.9 & 79.1 \Large $\pm$ 2.1 & 62.7 \Large $\pm$ 3.9 & 72.2 \Large $\pm$ 2.3  \\

\cmidrule(lr){1-22}

FedAvg+DART & 83.2 \Large $\pm$ 3.2 & 68.4 \Large $\pm$ 5.5 & 80.1 \Large $\pm$ 3.5 & 63.9 \Large $\pm$ 5.7 & 76.1 \Large $\pm$ 4.4 & 70.5 \Large $\pm$ 5.5 & 72.7 \Large $\pm$ 5.8 & 64.8 \Large $\pm$ 5.4 & 72.8 \Large $\pm$ 5.0 & 56.2 \Large $\pm$ 7.0 & 46.6 \Large $\pm$ 9.2 & 60.8 \Large $\pm$ 6.8 & 71.6 \Large $\pm$ 5.3 & 67.8 \Large $\pm$ 6.3 & 73.0 \Large $\pm$ 3.7 & 79.7 \Large $\pm$ 3.5 & 63.0 \Large $\pm$ 6.9 & 67.6 \Large $\pm$ 6.8 & 75.1 \Large $\pm$ 4.9 & 64.8 \Large $\pm$ 6.9 & 71.8 \Large $\pm$ 5.3  \\

FedDyn+DART & 79.1 \Large $\pm$ 1.5 & 67.2 \Large $\pm$ 1.6 & 77.1 \Large $\pm$ 1.7 & 63.1 \Large $\pm$ 3.3 & 74.4 \Large $\pm$ 2.5 & 71.1 \Large $\pm$ 2.3 & 70.6 \Large $\pm$ 3.2 & 63.5 \Large $\pm$ 3.2 & 72.2 \Large $\pm$ 2.8 & 56.2 \Large $\pm$ 3.7 & 48.8 \Large $\pm$ 3.6 & 59.5 \Large $\pm$ 0.9 & 71.8 \Large $\pm$ 1.4 & 67.9 \Large $\pm$ 3.0 & 69.7 \Large $\pm$ 3.2 & 75.9 \Large $\pm$ 2.3 & 62.2 \Large $\pm$ 3.1 & 65.0 \Large $\pm$ 1.8 & 72.3 \Large $\pm$ 1.8 & 63.2 \Large $\pm$ 2.8 & 72.4 \Large $\pm$ 2.5  \\

FedNova+DART & 80.3 \Large $\pm$ 0.7 & 67.6 \Large $\pm$ 1.1 & 77.4 \Large $\pm$ 1.0 & 59.4 \Large $\pm$ 1.9 & 74.6 \Large $\pm$ 1.1 & 71.1 \Large $\pm$ 1.5 & 69.1 \Large $\pm$ 1.4 & 63.6 \Large $\pm$ 1.6 & 71.9 \Large $\pm$ 1.5 & 58.2 \Large $\pm$ 1.5 & \textbf{50.1} \Large $\pm$ 1.6 & 60.3 \Large $\pm$ 1.6 & 72.7 \Large $\pm$ 1.0 & 67.1 \Large $\pm$ 1.8 & 72.1 \Large $\pm$ 1.6 & 77.4 \Large $\pm$ 0.6 & 63.8 \Large $\pm$ 1.2 & 65.8 \Large $\pm$ 1.1 & 73.8 \Large $\pm$ 0.9 & 64.7 \Large $\pm$ 1.2 & 71.3 \Large $\pm$ 1.9  \\

FedProx+DART & 87.1 \Large $\pm$ 0.9 & \textbf{73.9} \Large $\pm$ 1.5 & 84.4 \Large $\pm$ 0.7 & \textbf{71.0} \Large $\pm$ 1.8 & \textbf{82.0} \Large $\pm$ 1.2 & \textbf{77.4} \Large $\pm$ 1.3 & 78.4 \Large $\pm$ 1.6 & \textbf{70.6} \Large $\pm$ 1.7 & \textbf{79.1} \Large $\pm$ 1.3 & \textbf{60.8} \Large $\pm$ 3.3 & 49.7 \Large $\pm$ 2.4 & \textbf{67.4} \Large $\pm$ 2.0 & \textbf{77.2} \Large $\pm$ 2.7 & \textbf{74.8} \Large $\pm$ 2.4 & \textbf{75.9} \Large $\pm$ 3.8 & 83.8 \Large $\pm$ 0.9 & \textbf{68.5} \Large $\pm$ 2.8 & \textbf{72.9} \Large $\pm$ 1.4 & \textbf{80.7} \Large $\pm$ 1.4 & \textbf{70.5} \Large $\pm$ 2.8 & \textbf{78.6} \Large $\pm$ 1.3  \\

\midrule
\multicolumn{22}{c}{$\alpha_\text{IID} = 0.1$} \\
\midrule

FedAvg & 49.0 \Large $\pm$ 14.6 & 37.1 \Large $\pm$ 10.7 & 48.1 \Large $\pm$ 14.5 & 34.7 \Large $\pm$ 10.0 & 36.4 \Large $\pm$ 14.4 & 35.3 \Large $\pm$ 15.8 & 40.4 \Large $\pm$ 14.0 & 38.6 \Large $\pm$ 10.3 & 31.9 \Large $\pm$ 12.7 & 32.7 \Large $\pm$ 5.7 & 26.6 \Large $\pm$ 6.3 & 36.4 \Large $\pm$ 7.6 & 37.8 \Large $\pm$ 12.8 & 33.8 \Large $\pm$ 13.1 & 40.0 \Large $\pm$ 5.0 & 47.5 \Large $\pm$ 14.0 & 37.2 \Large $\pm$ 7.4 & 36.9 \Large $\pm$ 10.9 & 42.3 \Large $\pm$ 12.6 & 38.6 \Large $\pm$ 7.8 & 29.7 \Large $\pm$ 14.8  \\

FedDyn & 34.7 \Large $\pm$ 7.5 & 31.6 \Large $\pm$ 6.4 & 30.7 \Large $\pm$ 6.0 & 19.4 \Large $\pm$ 1.4 & 33.0 \Large $\pm$ 6.9 & 32.6 \Large $\pm$ 6.8 & 24.4 \Large $\pm$ 4.8 & 25.9 \Large $\pm$ 4.5 & 32.3 \Large $\pm$ 6.6 & 34.6 \Large $\pm$ 7.4 & 33.4 \Large $\pm$ 7.0 & 34.1 \Large $\pm$ 6.8 & 34.5 \Large $\pm$ 7.4 & 32.6 \Large $\pm$ 6.0 & 34.2 \Large $\pm$ 7.4 & 31.6 \Large $\pm$ 6.6 & 34.6 \Large $\pm$ 7.5 & 31.1 \Large $\pm$ 6.9 & 34.3 \Large $\pm$ 7.3 & 34.6 \Large $\pm$ 7.6 & 32.0 \Large $\pm$ 6.5  \\

FedNova & 54.3 \Large $\pm$ 3.8 & 45.0 \Large $\pm$ 3.1 & 50.5 \Large $\pm$ 4.5 & 26.1 \Large $\pm$ 5.8 & 44.0 \Large $\pm$ 3.5 & 44.3 \Large $\pm$ 4.0 & 34.7 \Large $\pm$ 7.5 & 43.4 \Large $\pm$ 4.8 & 40.6 \Large $\pm$ 3.4 & 48.4 \Large $\pm$ 1.3 & 43.9 \Large $\pm$ 2.4 & 45.2 \Large $\pm$ 0.9 & 52.5 \Large $\pm$ 3.3 & 37.9 \Large $\pm$ 3.1 & 52.1 \Large $\pm$ 3.7 & 51.1 \Large $\pm$ 3.7 & 50.2 \Large $\pm$ 1.9 & 48.8 \Large $\pm$ 2.8 & 51.6 \Large $\pm$ 2.8 & 50.3 \Large $\pm$ 2.2 & 40.1 \Large $\pm$ 3.2  \\

FedProx & \textbf{68.0} \Large $\pm$ 1.3 & 56.7 \Large $\pm$ 0.8 & \textbf{64.6} \Large $\pm$ 1.0 & 43.9 \Large $\pm$ 4.2 & 56.4 \Large $\pm$ 1.5 & 56.7 \Large $\pm$ 1.6 & 54.6 \Large $\pm$ 2.5 & 56.7 \Large $\pm$ 2.1 & 52.4 \Large $\pm$ 1.5 & 55.3 \Large $\pm$ 2.1 & \textbf{50.6} \Large $\pm$ 2.7 & 52.9 \Large $\pm$ 2.2 & \textbf{63.7} \Large $\pm$ 1.1 & 51.4 \Large $\pm$ 3.9 & \textbf{63.9} \Large $\pm$ 2.7 & \textbf{63.9} \Large $\pm$ 1.9 & 59.1 \Large $\pm$ 2.1 & \textbf{58.0} \Large $\pm$ 1.5 & \textbf{63.2} \Large $\pm$ 1.8 & 59.3 \Large $\pm$ 2.3 & 51.4 \Large $\pm$ 1.7  \\

\cmidrule(lr){1-22}

FedAvg+DART & 43.0 \Large $\pm$ 15.8 & 32.9 \Large $\pm$ 14.7 & 41.2 \Large $\pm$ 16.0 & 29.9 \Large $\pm$ 12.5 & 36.9 \Large $\pm$ 15.9 & 33.3 \Large $\pm$ 16.4 & 35.2 \Large $\pm$ 15.4 & 30.1 \Large $\pm$ 14.6 & 34.8 \Large $\pm$ 16.0 & 26.7 \Large $\pm$ 14.8 & 20.1 \Large $\pm$ 13.4 & 29.6 \Large $\pm$ 15.5 & 34.4 \Large $\pm$ 13.3 & 34.3 \Large $\pm$ 16.7 & 36.2 \Large $\pm$ 7.4 & 41.6 \Large $\pm$ 15.4 & 30.2 \Large $\pm$ 15.1 & 30.3 \Large $\pm$ 15.2 & 36.2 \Large $\pm$ 15.8 & 31.3 \Large $\pm$ 14.9 & 33.0 \Large $\pm$ 16.1  \\

FedDyn+DART & 23.8 \Large $\pm$ 15.3 & 21.9 \Large $\pm$ 13.6 & 21.1 \Large $\pm$ 13.9 & 16.0 \Large $\pm$ 7.6 & 22.9 \Large $\pm$ 14.7 & 22.6 \Large $\pm$ 14.3 & 18.8 \Large $\pm$ 10.8 & 18.2 \Large $\pm$ 11.4 & 22.5 \Large $\pm$ 14.4 & 23.1 \Large $\pm$ 14.4 & 22.6 \Large $\pm$ 14.1 & 22.7 \Large $\pm$ 13.8 & 23.6 \Large $\pm$ 15.2 & 22.2 \Large $\pm$ 13.5 & 23.3 \Large $\pm$ 15.0 & 22.0 \Large $\pm$ 13.9 & 23.3 \Large $\pm$ 14.7 & 21.4 \Large $\pm$ 14.1 & 23.2 \Large $\pm$ 14.7 & 23.3 \Large $\pm$ 14.7 & 22.6 \Large $\pm$ 14.6  \\

FedNova+DART & 48.0 \Large $\pm$ 6.0 & 40.5 \Large $\pm$ 5.1 & 43.4 \Large $\pm$ 6.7 & 26.2 \Large $\pm$ 5.9 & 42.0 \Large $\pm$ 5.7 & 40.8 \Large $\pm$ 5.9 & 32.7 \Large $\pm$ 7.3 & 35.7 \Large $\pm$ 5.8 & 40.0 \Large $\pm$ 5.3 & 42.7 \Large $\pm$ 4.0 & 37.3 \Large $\pm$ 4.0 & 40.3 \Large $\pm$ 4.0 & 46.0 \Large $\pm$ 5.6 & 37.7 \Large $\pm$ 4.7 & 45.6 \Large $\pm$ 5.8 & 44.8 \Large $\pm$ 6.0 & 44.0 \Large $\pm$ 4.5 & 40.5 \Large $\pm$ 5.0 & 44.9 \Large $\pm$ 5.3 & 44.0 \Large $\pm$ 4.7 & 40.0 \Large $\pm$ 5.6  \\

FedProx+DART & 66.9 \Large $\pm$ 4.3 & {\textbf{58.7}} \Large $\pm$ 2.1 &
63.9 \Large $\pm$ 2.6 & {\textbf{51.0}} \Large $\pm$ 0.0 & {\textbf{61.4}} \Large $\pm$ 5.3 &
{\textbf{59.3}} \Large $\pm$ 4.8 & {\textbf{58.3}} \Large $\pm$ 0.7 & {\textbf{57.1}} \Large $\pm$ 2.0 &
{\textbf{59.6}} \Large $\pm$ 5.2 & {\textbf{55.9}} \Large $\pm$ 1.6 & 49.8 \Large $\pm$ 0.0 &
{\textbf{53.5}} \Large $\pm$ 0.2 & 62.7 \Large $\pm$ 4.2 & {\textbf{59.1}} \Large $\pm$ 1.4 &
63.3 \Large $\pm$ 7.4 & 62.8 \Large $\pm$ 3.0 & {\textbf{59.4}} \Large $\pm$ 0.2 &
56.7 \Large $\pm$ 1.6 & 61.4 \Large $\pm$ 4.2 & {\textbf{60.1}} \Large $\pm$ 0.2 &
{\textbf{59.4}} \Large $\pm$ 4.2 \\

\bottomrule
\end{tabular}%
}}
\label{tab:full_plugin_result}
\end{table*}

\section{Ablation Studies}

\subsection{Generalization Across Model Architectures}
\label{appendix:models}
To demonstrate DART on various deep learning models, we conduct simulations for a 10 client FL setup with a fixed time budget of 1000s and we compare DART enhanced FedAvg clean, robust and average accuracy against FedAvg and FedAugMix for ResNet-18, MobileNet and VGG-16. Table~\ref{table:performance_results_iso-time} shows that DART enhances the robust accuracy of FedAvg by $4.3\%$ for a slight drop of $1.7\%$ in clean accuracy on average. Additionally, DART enhanced FedAvg outperforms FedAugMix in both clean and robust accuracy by an average of $18.9\%$ and $13.8\%$, respectively. Most notably, DART performs best with VGG-16 where the gains are most significant ($29.2\%$ in $\clnA$, $24.3\%$ in $ \robA$). These results indicate that DART-enhanced FL generalizes consistently across model architectures while maintaining a favorable balance between clean and robust accuracy, making it well-suited for deployment in resource-constrained environments.

\begin{table}[h]
\centering
\caption{Accuracy at iso-time of $\approx$ 1000s for different models. The value of $G$ is set to meet the time budget. FedAvg+DART achieves the highest robustness at zero client overhead.}
\label{table:performance_results_iso-time}
\resizebox{\linewidth}{!}{%
\Huge
\begin{tabular}{llcccc}
\toprule
Model                  & Method    & $G$ & $\clnA$ (\%) & $\robA$ (\%) & $\avgA$ (\%) \\
\midrule
\multirow{3}{*}{ResNet-18} 
                       & FedAvg   & 83    & \textbf{90.00}     & 71.32        & 80.66           \\
                       \cmidrule(lr){2-6}
                       & FedAugMix  & 27    & 72.22              & 65.97        & 69.10           \\
                       & FedAvg+DART    & 83    & 87.62              & \textbf{75.85} & \textbf{81.74} \\
\midrule
\multirow{3}{*}{MobileNet} 
                       & FedAvg   & 141   & \textbf{89.83}     & 72.38        & 81.11           \\
                       \cmidrule(lr){2-6}
                       & FedAugMix  & 43    & 76.30              & 68.41        & 72.36           \\
                       & FedAvg+DART    & 141   & 88.07              & \textbf{75.73} & \textbf{81.90} \\
\midrule
\multirow{3}{*}{VGG-16} 
                       & FedAvg   & 135   & \textbf{89.90}     & 72.08        & 80.99          \\
                       \cmidrule(lr){2-6}
                       & FedAugMix  & 36    & 59.31              & 52.83        & 56.07           \\
                       & FedAvg+DART    & 135   & 88.81              & \textbf{77.14} & \textbf{82.98} \\
\bottomrule
\end{tabular}
}
\end{table}

\begin{figure*}[h]
\begin{center}
    \includegraphics[width=\linewidth]{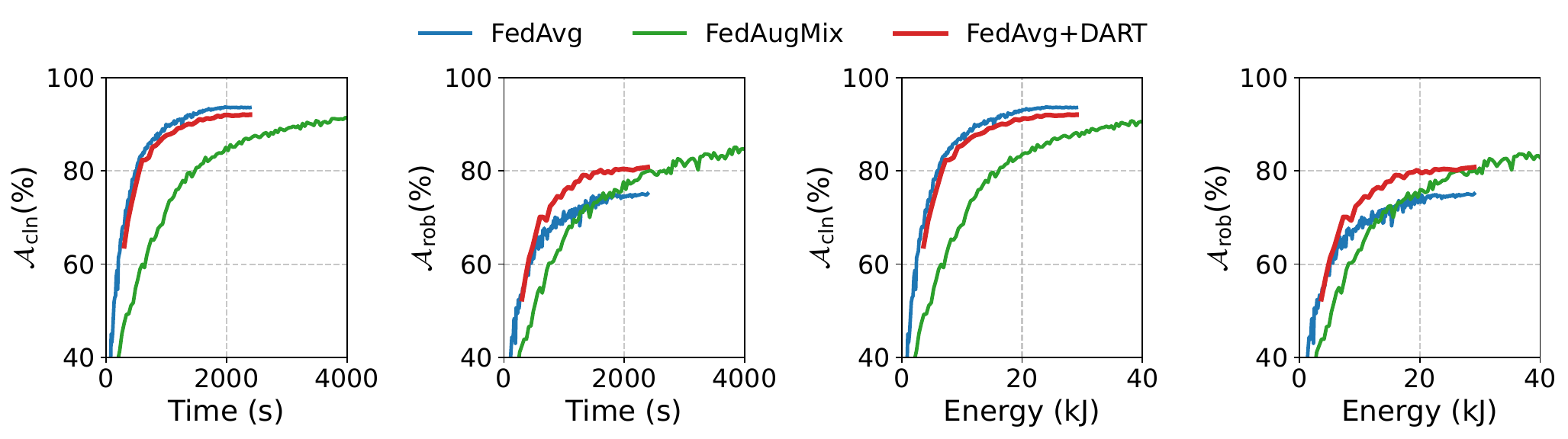}
    \caption{CIFAR-10 clean and CIFAR-10-C robust accuracy, $\clnA$ and $\robA$, under FedAvg, FedAugMix, and FedAvg+DART as a function of time and energy. The model used is ResNet-18, the server dataset is CIFAR-100, and 10 clients are deployed. FedAvg+DART takes the least energy and time to reach $\robA=80\%$.}
    \label{fig:curves_resnet}
\end{center}
\end{figure*}

\begin{figure*}[h]
\begin{center}
    \includegraphics[width=\linewidth]{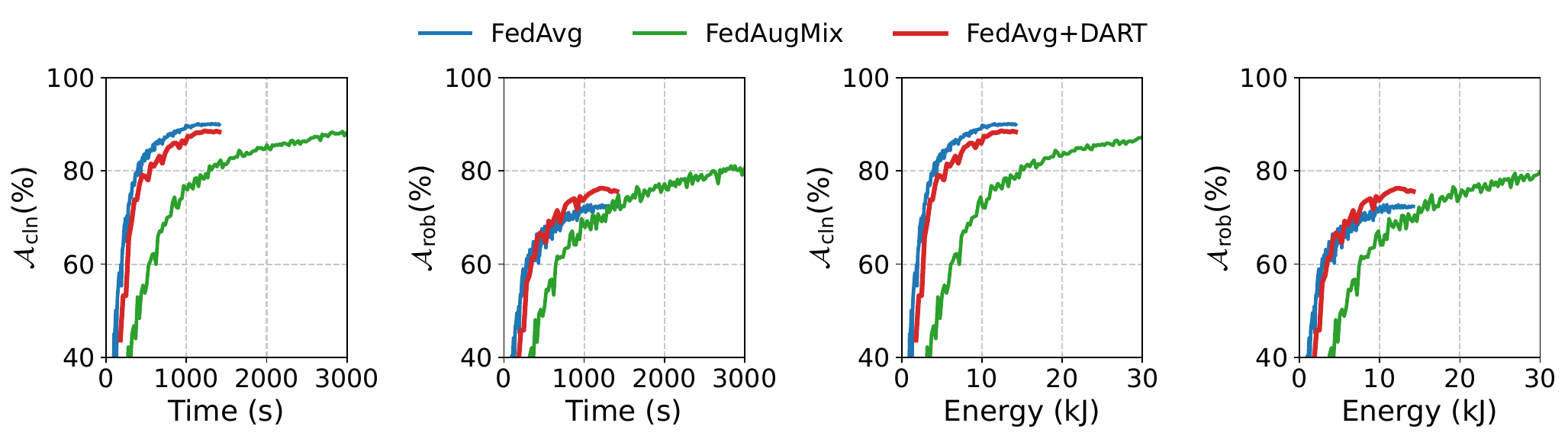}
    \caption{CIFAR-10 clean and CIFAR-10-C robust accuracy, $\clnA$ and $\robA$, under FedAvg, FedAugMix, and FedAvg+DART as a function of time and energy. The model used is MobileNet, the server dataset is CIFAR-100, and 10 clients are deployed.}
    \label{fig:curves_mobilenet}
\end{center}
\end{figure*}

\begin{figure*}[h]
\begin{center}
    \includegraphics[width=\linewidth]{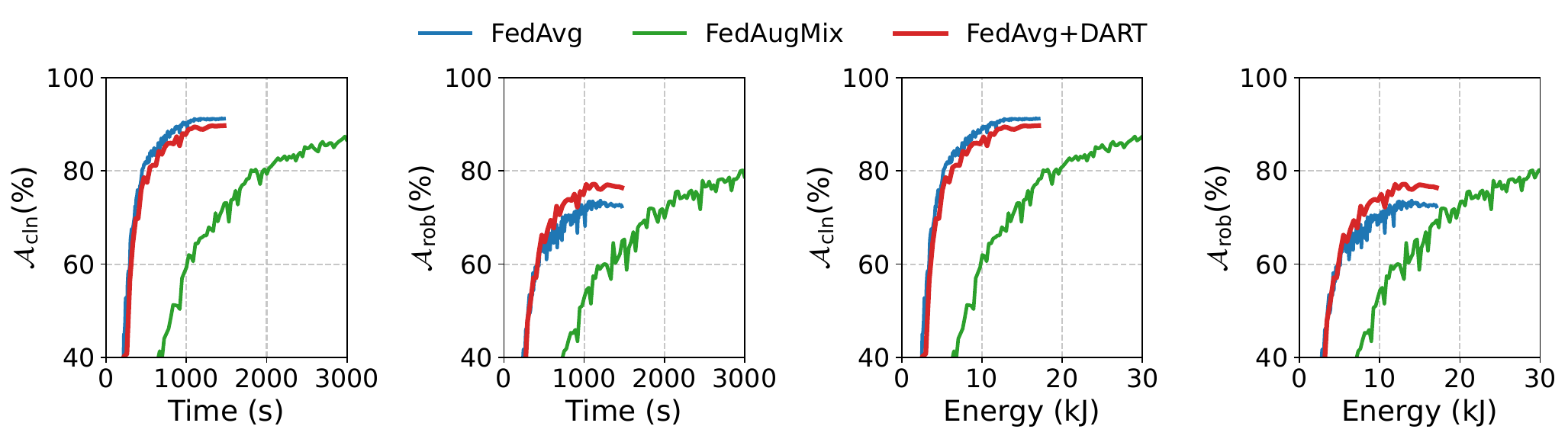}
    \caption{CIFAR-10 clean and CIFAR-10-C robust accuracy, $\clnA$ and $\robA$, under FedAvg, FedAugMix, and FedAvg+DART as a function of time and energy. The model used is VGG-16, the server dataset is CIFAR-100, and 10 clients are deployed.}
    \label{fig:curves_vgg}
\end{center}
\end{figure*}

Fig.~\ref{fig:curves_resnet} shows that DART-enhanced FedAvg consistently outperforms both FedAvg and FedAugMix in robust accuracy ($\robA)$ under constrained time and energy budgets. Across the full range of time and energy, DART-enhanced FL achieves higher robustness than FedAvg. When the per-client time budget is below 2200s or the energy budget is below 25kJ, DART also surpasses FedAugMix, highlighting its effectiveness in resource-constrained settings. Moreover, DART-enhanced FL reaches a target robust accuracy of 80\% on ResNet-18 with the lowest time and energy cost among all methods. In contrast, FedAugMix requires substantially more computation to converge, limiting its practicality for deployment on resource-constrained devices. All methods are evaluated with 10 clients over 200 global rounds, with DART applied to the FedAvg model at each round. Similar trends are observed for MobileNet (Fig.~\ref{fig:curves_mobilenet}) and VGG-16 (Fig.~\ref{fig:curves_vgg}), demonstrating DART’s generalization across model architectures.

\subsection{CIFAR-100 Results}

\label{appendix:c100}

To showcase the benefit of using DART with different client datasets, we provide the following results that utilize CIFAR-100 on the clients and CIFAR-10 on the server. The number of clients used is 10.

\begin{figure}[h]
\centering
\begin{subfigure}[b]{0.49\linewidth}
    \includegraphics[width=\linewidth]{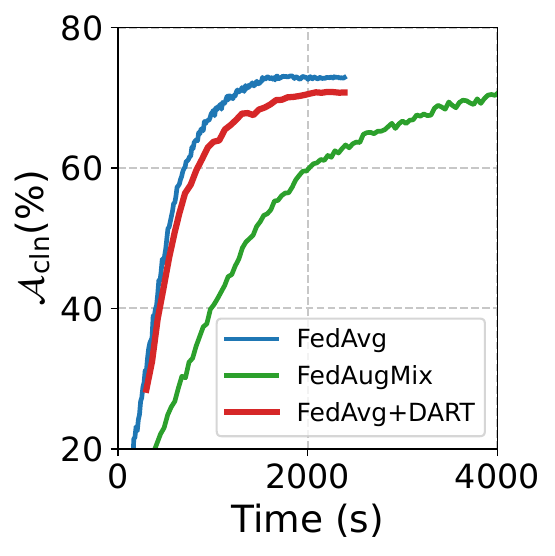}
\end{subfigure}
\begin{subfigure}[b]{0.49\linewidth}
    \includegraphics[width=\linewidth]{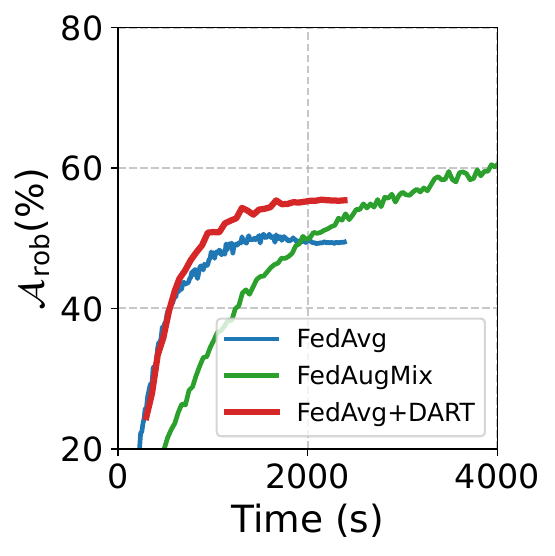}
\end{subfigure}
\begin{subfigure}[b]{0.49\linewidth}
    \includegraphics[width=\linewidth]{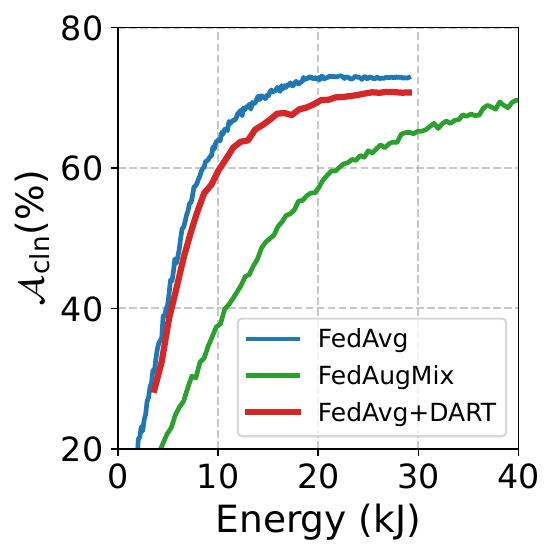}
\end{subfigure}
\begin{subfigure}[b]{0.49\linewidth}
    \includegraphics[width=\linewidth]{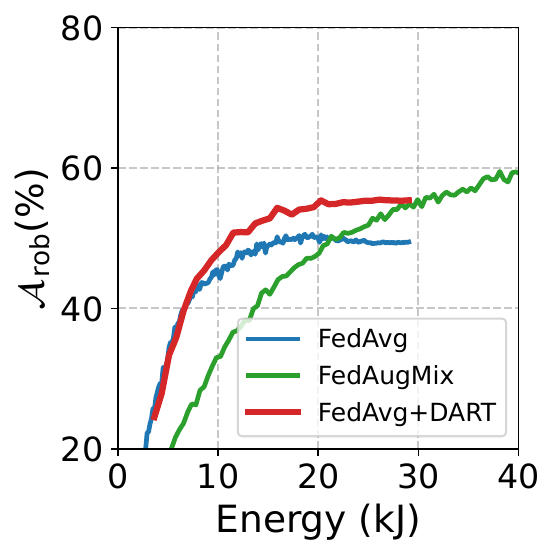}
\end{subfigure}
\caption{CIFAR-100 clean and CIFAR-100-C robust accuracy, $\clnA$ and $\robA$, under FedAvg, FedAugMix, and FedAvg+DART as a function of time and energy. The model used is ResNet-18, the server dataset is CIFAR-10, and 10 clients are deployed.}
\label{fig:curves_resnet_100}
\end{figure}

Analyzing the CIFAR-100 training curves of ResNet-18 in Fig.~\ref{fig:curves_resnet_100}, we notice that DART enhanced FedAvg achieves the best robust accuracy when time and energy are below 2500s and 30kJ, respectively. Within these intervals, DART also significantly outperforms FedAugMix in clean accuracy, while incurring a drop relative to FedAvg. As expected, FedAugMix requires substantially more time and energy to converge. DART provides the most robust model under resource constraints, even when CIFAR-100 is the client dataset.

These experiments demonstrate that the advantages of DART are not limited to a specific client dataset. By maintaining strong clean and robust performance across diverse data distributions, DART proves to be broadly applicable and adaptable, reinforcing its potential as a versatile solution for robust federated learning in scenarios constrained by compute, memory, time, and energy.

\subsection{Impact of Robustification Period}

We experimentally investigate the impact of the robustness-enhancement period $\robT$ on utility and robustness. Additional DART iterations can be executed since the server is not resource-limited and this imposes no further cost on clients.

\begin{figure}[h]
\begin{center}
    \includegraphics[width=\linewidth]{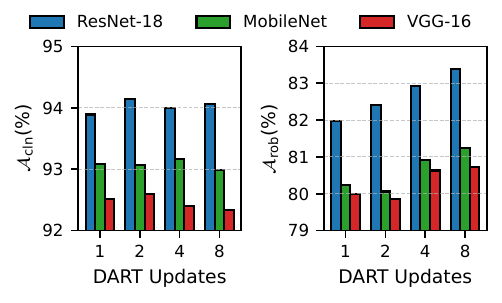}
    \caption{FedAvg+DART CIFAR-10 clean and CIFAR-10-C robust accuracy for different numbers of DART updates. The number of global updates, $E$, is set to 200. The number of clients is 4 and the server dataset is CIFAR-100.}
    \label{fig:trob_sweep}
\end{center}
\end{figure}

Fig.~\ref{fig:trob_sweep} shows that reducing the robustification period $\robT$ has minimal impact on clean accuracy. In general, smaller values of $\robT$, i.e., more frequent DART updates, leads to higher robust accuracy at no additional client training time and energy. For example, ResNet-18 robust accuracy improves from $81.98\%$ to $83.39\%$ when $\robT$ decreases from 200 to 25. Similar trends can be observed for MobileNet and VGG-16, where robust accuracy increases by $+1\%$ and $+0.72\%$ respectively. These findings suggest that robust accuracy can be increased by performing more frequent DART updates, however, in cases where the server cannot be overloaded, applying DART once at the final global update yields competitive performance.

\subsection{DART Parameter Sweeps}
\label{appendix:alpha}
\begin{figure}[h]
\begin{center}
    \includegraphics[width=\linewidth]{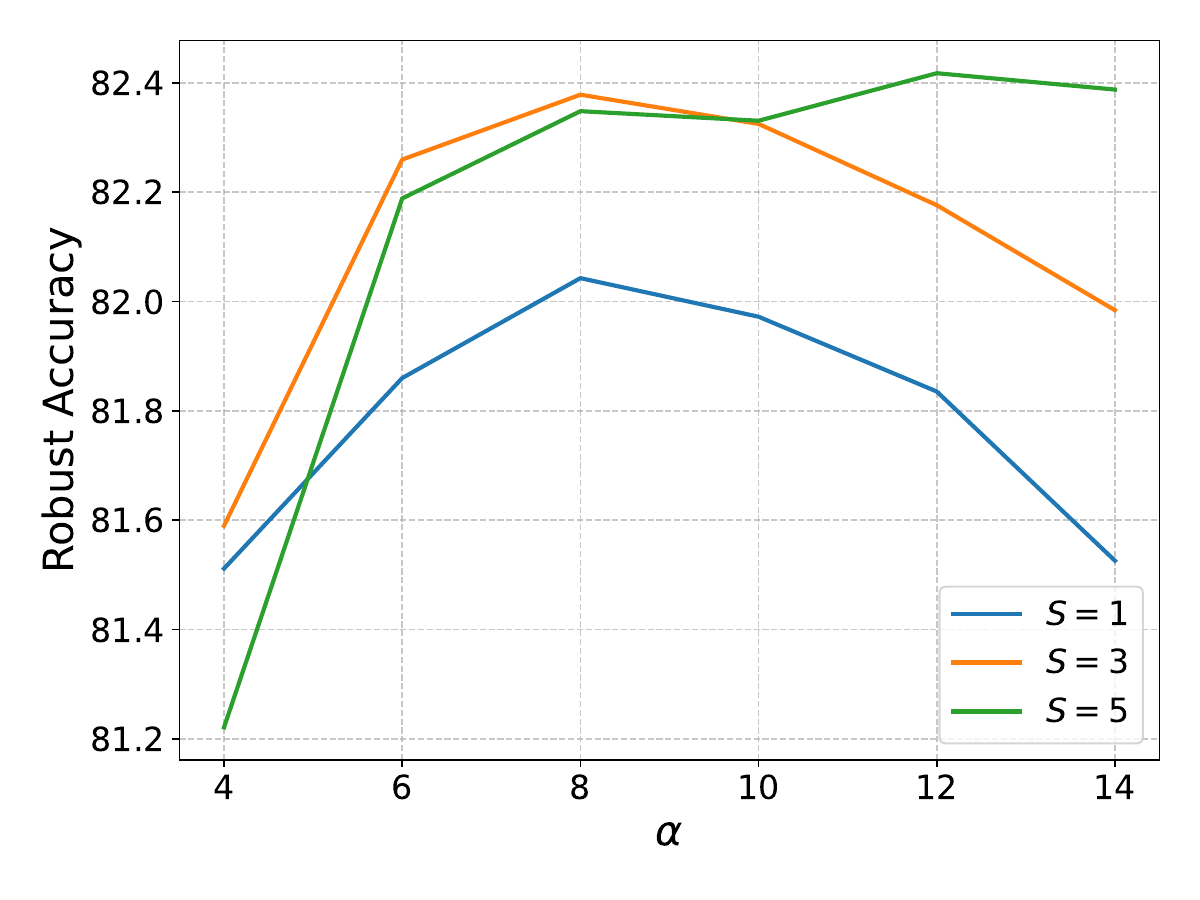}
    \caption{Impact of DART loss weighing parameter $\alpha$ and data augmentation mixture width $S$ on CIFAR-10-C robust accuracy. We choose $\alpha=10$ and $S=3$ to maximize robustness and limit the server workload.}
    \label{fig:sweep}
\end{center}
\end{figure}

In DART, performance is sensitive to the loss-weighting parameter \(\alpha\) and the mixture width \(S\). Fig~\ref{fig:sweep} reports the effect of varying \(\alpha\) and \(S\) on robust accuracy. To maximize robustness, we observe that \(\alpha\) values in the range of 8--10 yield the strongest performance. Increasing \(S\) introduces more complex data augmentation but also increases server-side computation. Based on this trade-off, we select \(\alpha = 10\) and \(S = 3\) to achieve strong robustness while limiting server overhead.

\subsection{Loss Component Ablations}
Table~\ref{table:loss} presents an ablation study on the components $\closs$ and $\dloss$ of the DART loss \eqref{eq:DARTloss} as compared to standard clean training. Applying DART without the consistency loss $\closs$ results in a model with clean and robust accuracy comparable to the original pre-trained model. In contrast, removing the distillation loss $\dloss$ leads to significant drops in both utility and robustness. Only when both loss components are present does DART maintain clean accuracy (with a minor $1.5\%$ drop) while improving robustness by $5.7\%$. 

These findings are consistent with those in Section~\ref{sec:proposed_method}: $\dloss$ preserves clean accuracy, while $\closs$ promotes consistency, which translates to improved robust accuracy with minimal loss in clean accuracy, when both loss components are jointly optimized.

\begin{table}[h]
\centering
\caption{CIFAR-10 clean and CIFAR-10-C robust accuracy results for FedAvgand FedAvg+DART with custom DART. The model used is ResNet-18. To improve robustness, DART must include both the consistency loss $\closs$ and the distillation loss $\dloss$.}
\label{table:loss}
\resizebox{0.5\textwidth}{!}{%
\tiny
\begin{tabular}{lccc}
\toprule
Method         & $\clnA$ (\%) & $\robA$ (\%) & $\avgA$ (\%) \\
\midrule
Clean training     & \textbf{93.58} & 75.09 & 84.34 \\
DART w/o $\closs$  & 93.46 & 75.36 & 84.41 \\
DART w/o $\dloss$  & 19.24 & 17.53 & 18.39 \\
DART               & 92.08 & \textbf{80.79} & \textbf{86.44} \\
\bottomrule
\end{tabular}
}
\end{table}

\begin{figure}[h]
    \centering
    \begin{subfigure}{0.32\linewidth}
        \includegraphics[width=\linewidth]{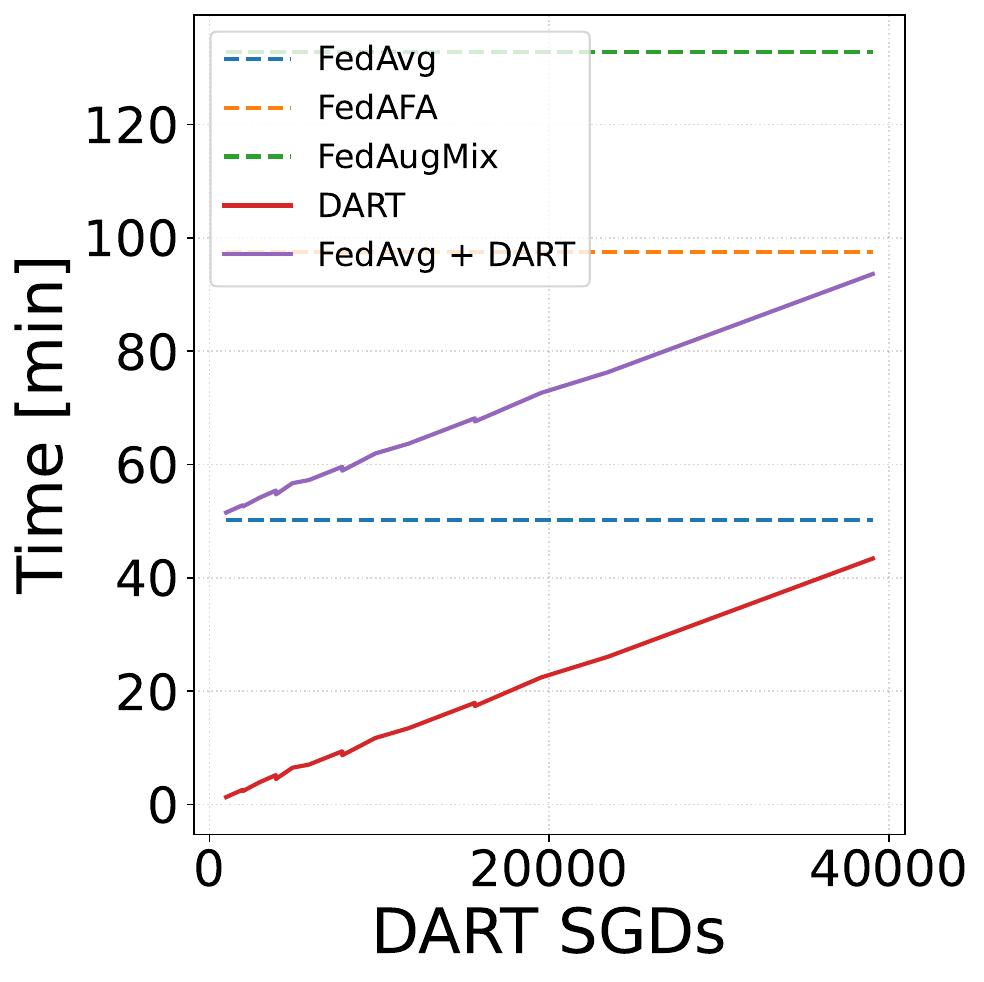}
        \caption{1 GPU}
    \end{subfigure}
    \hfill
    \begin{subfigure}{0.32\linewidth}
        \includegraphics[width=\linewidth]{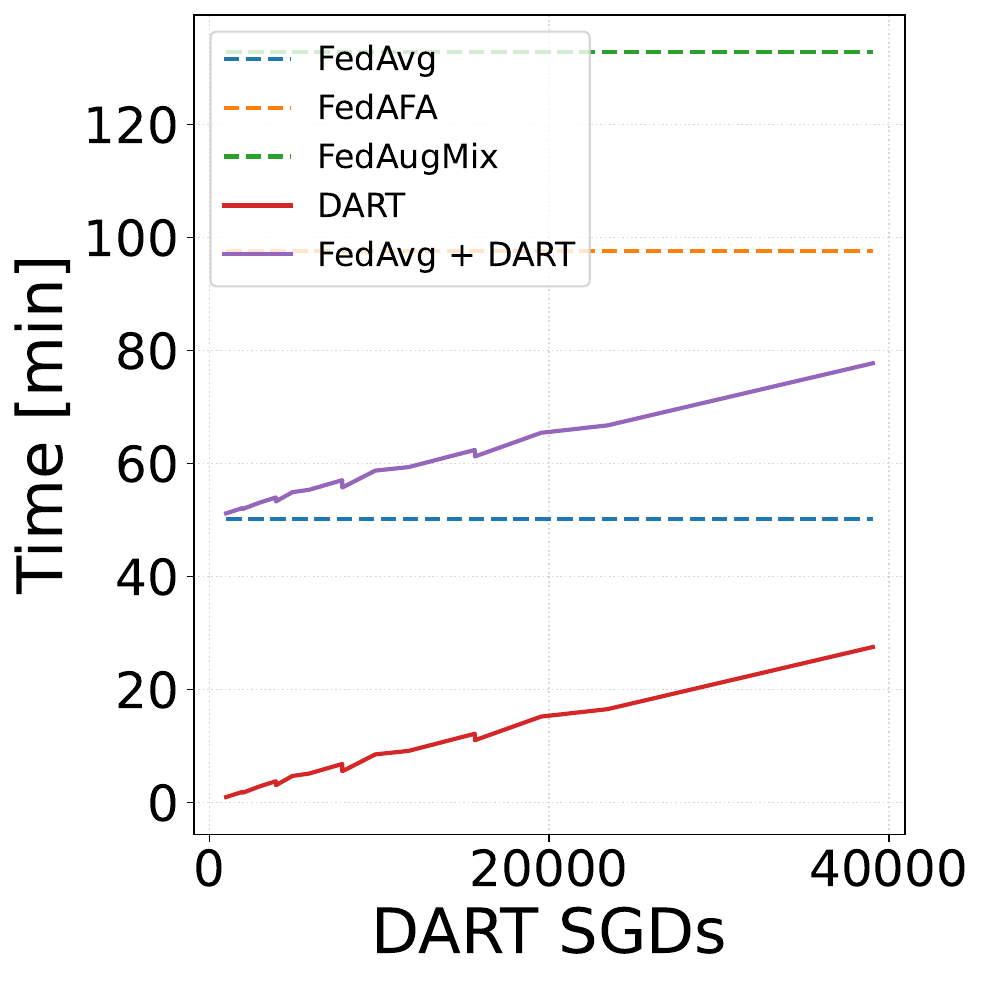}
        \caption{2 GPUs}
    \end{subfigure}
    \hfill
    \begin{subfigure}{0.32\linewidth}
        \includegraphics[width=\linewidth]{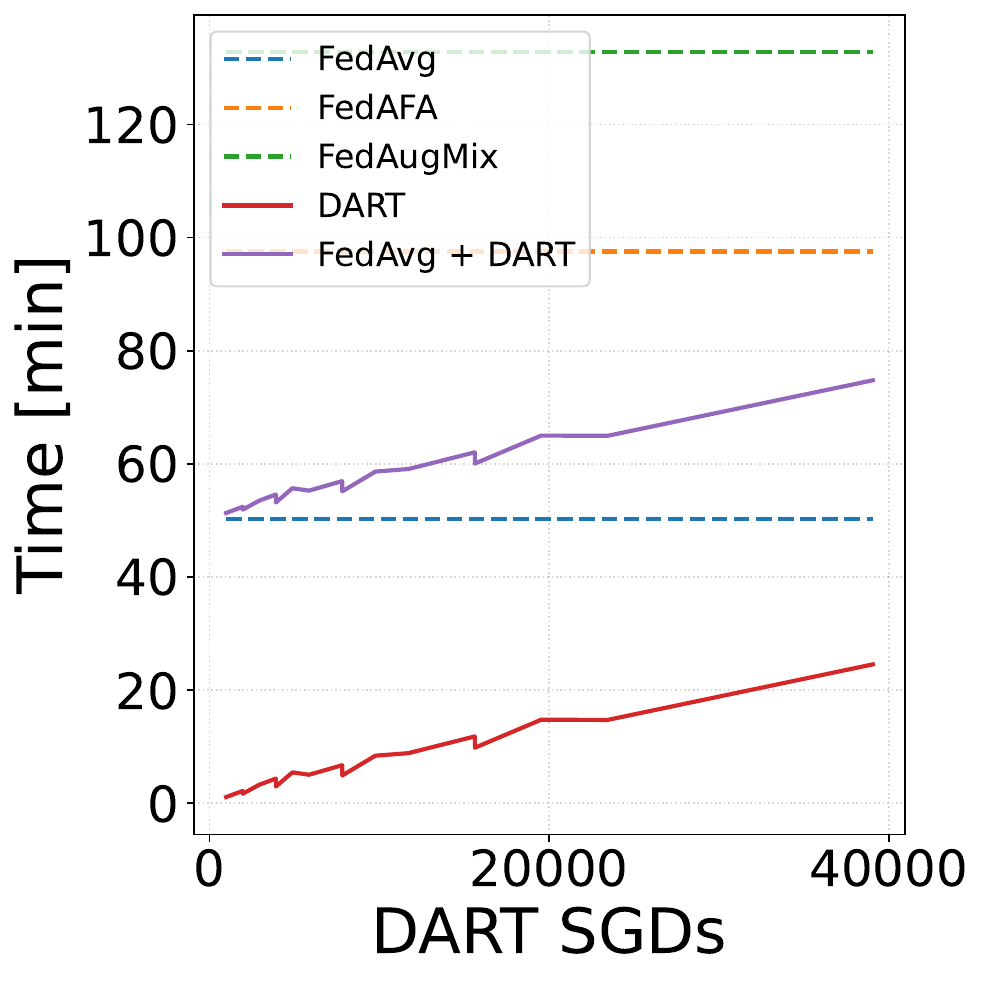}
        \caption{4 GPUs}
    \end{subfigure}
    \caption{Total time versus number of DART SGD updates. DART-enhanced FedAvg maintains lower training time compared to robust FL methods.}
    \label{fig:time_sgd}
\end{figure}
\vskip -0.1in

\begin{table}[h]
\centering
\caption{Impact of DART on CIFAR-10 clean and robust accuracy with BigGAN-generated server data.}
\label{tab:syn}
\resizebox{0.7\linewidth}{!}{%
\begin{tabular}{lccc}
\toprule
Method & $\alpha_\text{iid}$ & $\clnA$ (\%) & $\robA$ (\%) \\
\midrule
FedAvg     & 100 &  89.8 & 70.8 \\
FedAvg     & 10 & 89.5 & 70.1 \\
\midrule
DART+FedAvg  & 100 & 86.4 & 73.7 \\
DART+FedAvg  & 10 & 86.3 & 74.9 \\

\bottomrule
\end{tabular}%
}
\end{table}

\section{Using Synthetic Data on the Server}
\label{appendix:gen}

When high-quality server data is unavailable, generative models can be used to supply the server dataset for DART. Table~\ref{tab:syn} shows that using BigGAN~\cite{brock2018large}-generated data with FedAvg+DART still yields substantial robustness gains, improving $\robA$ by 4.8

\section{Accounting for Server-side Costs in DART-enhanced FL}
\label{appendix:server}

DART offloads the resource intensive robust training to the server. Fig.~\ref{fig:time_sgd} shows that the large server side resources and parallelism reduce overall training time when DART is applied to FedAvg compared to other robust FL works. Crucially, client-side time and energy remain comparable to non-robust methods as shown in~\ref{tab:robust_efficiency}.